\documentclass[twoside]{article}

\usepackage[accepted]{aistats2025}
\usepackage[round]{natbib}

\usepackage{amsmath}
\usepackage{amssymb}
\usepackage{mathtools}
\usepackage{amsthm}
\usepackage{times}
\usepackage{latexsym}
\usepackage{xspace,mfirstuc,tabulary}
\usepackage{enumitem}
\usepackage{pifont}% http://ctan.org/pkg/pifont
\usepackage{graphicx}
\usepackage{amsmath}
\usepackage{numprint}
\usepackage{booktabs}
\usepackage{multirow}
\usepackage{lipsum}
\usepackage{amssymb}
\usepackage[hidelinks]{hyperref}
\usepackage{breakcites}
\usepackage{xcolor}
\usepackage{bbold}

\usepackage{stfloats}
\usepackage[makeroom]{cancel}
\usepackage{kantlipsum}
\usepackage{MnSymbol}
\usepackage{float}
\usepackage{subfig}

\allowdisplaybreaks

\usepackage[capitalize,noabbrev]{cleveref}

\theoremstyle{plain}
\newtheorem{theorem}{Theorem}[section]
\newtheorem{proposition}[theorem]{Proposition}
\newtheorem{lemma}[theorem]{Lemma}
\newtheorem{corollary}[theorem]{Corollary}
\theoremstyle{definition}

\newtheorem{assumption}[theorem]{Assumption}
\theoremstyle{remark}

\newcommand{\X}{\mathcal{X}}
\newcommand{\A}{\mathcal{A}}
\newcommand{\R}{\mathbb{R}}
\newcommand{\C}{\mathcal{C}}

\newcommand{\D}{\mathcal{D}}
\newcommand{\Exp}[2]{
  \ensuremath{\mathbb{E}_{#1}\left[#2\right]}}

\newcommand{\Var}[2]{
  \ensuremath{\mathbb{V}_{#1}\left[#2\right]}}

\begin{document}


%\runningauthor{Surname 1, Surname 2, Surname 3, Surname n}

\twocolumn[

\aistatstitle{Clustering Context in Off-Policy Evaluation}

\aistatsauthor{ 
Daniel Guzmán-Olivares \And 
Philipp Schmidt \And 
Jacek Golebiowski\And 
Artur Bekasov
}

\aistatsaddress{
Bulil Technologies, UAM\textsuperscript{*} \\
\href{mailto:dgo@buliltec.com}{\small daniel.guzman@buliltec.com} 
\And 
Amazon \\
\href{mailto:phschmid@amazon.de}{\small phschmid@amazon.com}
\And 
distil labs\textsuperscript{*} \\
\href{mailto:golebiowski.j@gmail.com}{\small golebiowski.j@gmail.com}
\And 
Amazon \\
\href{mailto:abksv@amazon.com}{\small abksv@amazon.com}
}
]

% Use '*' for this specific footnote
\renewcommand{\thefootnote}{*}
\footnotetext{Work done while at Amazon.}

% Reset footnote counter and numbering for the main text
\setcounter{footnote}{0}
\renewcommand{\thefootnote}{\arabic{footnote}}

\begin{abstract}
Off-policy evaluation can leverage logged data to estimate the effectiveness of new policies in e-commerce, search engines, media streaming services, or automatic diagnostic tools in healthcare. However, the performance of baseline off-policy estimators like IPS deteriorates when
the logging policy significantly differs from the evaluation policy. Recent work proposes sharing information across similar actions to mitigate this problem. 
In this work, we propose an alternative estimator that shares information across similar contexts using clustering.
We study the theoretical properties of the proposed estimator, characterizing its bias and variance under different conditions. 
We also compare the performance of the proposed estimator and existing approaches in various synthetic problems, as well as a real-world recommendation dataset.
Our experimental results confirm that clustering contexts improves estimation accuracy, especially in deficient information settings.\footnote{The code for reproducing our experimental implementation is available at \href{https://github.com/amazon-science/ope-cluster-context}{https://github.com/amazon-science/ope-cluster-context}}
\end{abstract}

\section{INTRODUCTION}
\label{section:1}
The contextual bandit process models many real-world problems across industry and research, including healthcare, finance, and recommendation systems \citep{survey}. In this setting, an agent observes a \textit{context}, chooses an action according to a \textit{policy}, and observes a \textit{reward}. \textit{Off-policy evaluation} (OPE) methods aim to estimate the effectiveness of a policy without empirically testing it, which can be particularly useful when A/B tests are costly, or if there is an inherent risk associated with poor policy performance, as is often the case in healthcare \citep{bastani}. Existing OPE methods can be broadly divided into parametric methods based on the \textit{direct method} (DM), non-parametric methods based on \textit{inverse propensity score} weighting \citep[IPS,][]{ips}, and a combination of the two, such as the \textit{doubly robust} method \citep[DR,][]{dr}. 
When every action with non-zero probability under the evaluation policy also has a non-zero probability under the logging policy, IPS is unbiased. 
This condition is rarely satisfied in real-world problems, however, so IPS is typically biased in practice, especially for actions that violate the condition, or have close-to-zero probabilities in the logging policy \citep{Sachdeva2020, dr, mips}. 
% In whis work we focus on the IPS-based approach to the OPE problem. In particular, our starting point is the 

Recently proposed \textit{Marginalized Inverse Propensity Score} estimator \citep[MIPS,][]{mips} improves upon IPS in large action spaces by pooling information across \textit{action embeddings}.
At the same time, MIPS suffers from the same problem as IPS for contexts in which a significant proportion of actions have low probability under the logging policy. In this case, MIPS lacks information about the actions to accurately estimate the importance weights, resulting in additional bias. In our work, we hypothesize that closeness at the context level should translate into similar behaviour for actions and rewards (for example, two movies of the same franchise in a recommendation system).
Based on this hypothesis, we propose an estimator that \emph{clusters} the context space, and pools information across all the contexts within a cluster.  %to make a more precise estimation of the importance weights. 
Informally, the proposed method solves the problem of deficient action information for a particular context by leveraging the information from all other contexts within the same cluster. 

We define and analyze the theoretical bandit setup with context clusters in Section \ref{section:3}, which leads to the formal derivation of the CHIPS estimator, for which we analyze bias and variance. In section \ref{section:4}, we compare the estimator's performance to the baselines on several synthetic and real-world datasets, verifying the theoretical findings, and demonstrating its effectiveness. Finally section \ref{section:5} explores future lines of work and CHIPS' limitations.

\section{BACKGROUND ON OFF-POLICY EVALUATION AND RELATED WORK}
\label{section:2}
The off-policy evaluation problem (OPE) is usually framed inside the general contextual bandit setup. Given an agent, determined by the policy $\pi: \X \times \A \rightarrow [0,1]$, the bandit's data generation process is defined as iterative logging of the agent's behavior when presented with different contexts. In each iteration, a context $x \in \X \subseteq \R^{d_x}$ is drawn i.i.d. from an unknown probability distribution $p(x)$ over the context space, an action $a \sim \pi(a|x)$ is selected from a finite action space $\A$, and a bounded reward $r \in [0, R_{\text{max}}]$ is observed as a sample from an unknown conditional distribution $p(r |a, x)$. The off-policy evaluation problem has been extensively studied from both a theoretical \citep{McNellisEOP17, counterfactual, Dumitrascu, irpan, switch} and a practical point of view given its applications in fields such as recommendation systems \citep{li, bendada, obd} or healthcare \citep{Varatharajah}.

We measure the performance of a policy $\pi$ through its \emph{value}, that we define as:
\begin{equation}
    \label{policy-value}
    V(\pi) := \mathbb{E}_{p(x)\pi(a|x)p(r|a,x)}[r] = \mathbb{E}_{p(x)\pi(a|x)}[q(a,x)]
\end{equation}
Here $q(a,x) = \Exp{p(r | a, x)}{r}$ denotes the conditional expected reward given an action $a$ and a context $x$.

In practice, we are interested in finding a policy maximizing the expected reward observed in the bandit process. A vital part of this process is the off-policy evaluation problem, in which we estimate the value of a policy $\pi$ given a dataset $\mathcal{D}:=\{(x_i, a_i, r_i)\}_{i=1}^N$ collected under a logging policy $\pi_0$ (i.e. $\mathcal{D} \sim \prod_{i=1}^N p(x)\pi_0(a|x)p(r |a, x)$). We use the mean squared error (MSE) to quantify how well the estimate $\hat{V}(\pi)$ approximates the real policy value $V(\pi)$:
\begin{align*}
    \text{MSE}(\hat{V}) &= \Exp{\mathcal{D}}{\left(V(\pi) - \hat{V}(\pi;\mathcal{D})^2\right)} \\
    &= \text{Bias}\left(\hat{V}(\pi;\mathcal{D})\right)^2 + \Var{\mathcal{D}}{\hat{V}(\pi;\mathcal{D})}
\end{align*}
A wide variety of approaches have been proposed in the literature to estimate $V(\pi)$. From them, three can be distinguished for being commonly used as starting points for developing new estimators. The first one is the Direct Method (DM), which tries to estimate $q(a, x)$ directly from \cref{policy-value}:
\begin{equation*}
  \hat{V}_{\text{DM}}(\pi;\mathcal{D}, \hat{q}) = \frac{1}{N}\sum_{i=1}^N\sum_{a \in \A}\hat{q}(a, x_i)
\end{equation*}
The bias of DM depends on the accuracy of the $\hat{q}(a, x) \approx q(a, x)$ approximation, but the variance is usually lower than in other approaches. Supervised learning in the DM's approach can be particularly useful when generalization of an agent's behaviour is needed due to limited information in the logging data \citep{Sachdeva2020}. However, when the reward function has a high variance, or the representation capacity is limited for the context-action pairs in the evaluation policy domain, $\hat{q}(a, x)$ could fail to accurately approximate $q(a,x)$ \citep{mrdr, beygelzimer, kallus}. This problem, known as \textit{reward misspecification}, can be quite difficult to detect in real-world examples \citep{mrdr, voloshin}, and is the reason why DM is generally regarded as a highly biased estimator. 

The second base approach is Inverse Propensity Scoring \citep[IPS,][]{ips}, which approximates the policy value by reweighting the rewards to correct the shift in action probabilities between the logging and evaluation policies:
\begin{align*}
  \hat{V}_{\text{IPS}}(\pi;\mathcal{D}) &= \frac{1}{N}\sum_{i=1}^N\frac{\pi(a_i|x_i)}{\pi_0(a_i|x_i)}r_i =\frac{1}{N}\sum_{i=1}^N w(a_i, x_i) r_i
\end{align*}
As per this definition, the context-action pairs selected by $\pi$ in which $\pi_0(a|x) = 0$ could be problematic, which motivates the following assumption:
\begin{assumption}(\textit{Common Support}) 
\label{as1}
Given an evaluation policy $\pi$ and a logging policy $\pi_0$, the latest has common support for $\pi$ if
    $$\pi_0(a | x) > 0\quad \forall a\in\A, x\in\X\mkern5mu:\mkern5mu \pi(a | x) > 0$$
\end{assumption}
The IPS estimator is unbiased under \cref{as1}. However, even when assumption \ref{as1} holds, IPS can present excessive variance due to the weights $w(a_i,x_i)$ taking larger values \citep{dr, mips}. This case is especially notable when $\pi_0$ and $\pi$ are significantly different or when trying to achieve universal support ($\pi_0(a|x) > 0 \mkern5mu\forall a\in\A, x \in \X$) in large action spaces \citep{mips, action-clusters, counterfactual}. Controlling the scaling of the propensity scores has motivated many approaches based on IPS, using techniques such as weight clipping \citep{dros, slope, clips} and self normalization \citep{snips, Kuzborskij2020ConfidentOE}. 
The Doubly Robust (DR) estimator combines DM and IPS, aiming to obtain a low-bias, low-variance estimate:
\begin{align*}
    V_{\text{DR}}(\pi; \mathcal{D}, \hat{q}) &:= V_{\text{DM}}(\pi; \hat{q}) \\
    &\quad + \frac{1}{N} \sum_{i=1}^N w(a_i, x_i)(r_i - \hat{q}(a_i, x_i))
\end{align*}
The DR estimator has been the cornerstone of multiple approaches that modify the base estimator to address problems such as low overlap between $\pi$ and $\pi_0$ \citep{switch, dr-lambda, Zhan2021OffPolicyEV, Guo2024DistributionallyRP}, reward misspecification \citep{mrdr}, and limited samples in logging data \citep{dros, Felicioni2022OffPolicyEW}. Unfortunately, the DR estimator can still inherit the large variance problem from IPS, for example, when dealing with large action spaces \citep{conjunct, mips, shimizu2023doubly, sachdeva2023offpolicy, taufiq2023marginal}. The problem of dealing with large action spaces was recently studied, resulting in the \textit{Marginalized Inverse Propensity Scoring} (MIPS) \citep{mips} estimator, in which the authors pool information between similar actions given some embedding representation $e \in \mathcal{E} \subset \mathbb{R}^d_e$ of them to address deficient actions in the logging policy. For this purpose, they introduce an IPS-based estimator marginalizing the probability over the action space:
\begin{align}
\hat{V}_{\mathrm{MIPS}}(\pi ; \mathcal{D}) & := \frac{1}{n} \sum_{i=1}^n \frac{p\left(e_i \mid x_i, \pi\right)}{p\left(e_i \mid x_i, \pi_0\right)} r_i \nonumber \\
& = \frac{1}{n} \sum_{i=1}^n w\left(x_i, e_i\right) r_i.
\end{align}
Where $p(e| x, \pi) := \sum_{a\in \mathcal{A}} p(e| x, a) \pi(a| x)
$.
The idea of estimating deficient items' behaviour by \textit{closely} observed ones inspired new approaches, like partitioning the action space in clusters \citep{action-clusters, conjunct}, or an adaptive method for ranking policies by optimizing user classification into given behavioural models and estimating independently for each group \citep{clutser-users-ranking}. The MR estimator \citep{taufiq2023marginal} diverged from the action space transformations and proposed marginalization over the rewards density through a regression estimate of the importance weights:
\begin{equation}
\hat{V}_{\mathrm{MR}}(\pi ; \mathcal{D}):=\frac{1}{n} \sum_{i=1}^n w\left(r_i\right) r_i
\end{equation}

Where $w(r)$ is defined as:
\begin{equation}
\begin{aligned}
\label{eq:mr}
w(r) := f_{\phi^*}(r):= & \operatorname{argmin} \mathbb{E}_\phi\left[\left(w(a, x)-f_\phi(r)\right)^2\right] \\
& f_\phi \in\left\{f_\phi: \mathbb{R} \rightarrow \mathbb{R} \mid \phi \in \Phi\right\}
\end{aligned}
\end{equation}

Motivated by these approaches, as well as the fact that estimating from \textit{similar} actions or make a regression over rewards could prove challenging if a significant proportion of these actions are missing for a given context, we propose the \textit{Context-Huddling Inverse Propensity Score} (CHIPS) estimator that we introduce in the next section.

\section{THE CHIPS ESTIMATOR}
\label{section:3}
The CHIPS estimator is based on the idea of partitioning the context space into clusters to extrapolate the behaviour of an agent when presented with a previously unseen or underrepresented context $x$. The assumption needed for this approximation to the OPE problem is that, given a policy, all contexts belonging to a cluster $c$ should have a similar probability of observing an action $a$ and will observe similar rewards when that action is chosen. Formally, we will consider a finite partition of the context space as the cluster space $\C := \{\C_i\}_{i=1}^K$ with $\C_i \subset \X$ and $c_i \cap \C_j = \emptyset$. We assume that we are given a $c \in \C$ for each context $x \in \X$, where we assume that $c$ is drawn i.i.d from an unknown distribution $p(c|x)$. Thus, given a policy $\pi$, we can compute its value by refining \cref{policy-value}:
\begin{align}
    V(\pi) & := \mathbb{E}_{p(x)p(c|x)\pi(a|x)p(r|a, c, x)}[r] \nonumber \\
    & = \mathbb{E}_{p(x)p(c|x)\pi(a|x)}[q(a, c, x)].
\end{align}

Where we denote $q(a, c, x) := \Exp{p(r|a, c, x)}{r}$ and it is important to note that $\Exp{p(c|x)\pi(a|x)}{q(a, c, x)} = \Exp{\pi(a|x)}{q(a,x)}$, and therefore the refinement is consistent with \cref{policy-value}. Similar to the common support condition in IPS, we formulate the following property as the equivalent for the CHIPS estimator of \cref{as1}.

\begin{assumption}(\textit{Common Cluster Support})
\label{as2}
Given an evaluation policy $\pi$ and a logging policy $\pi_0$, the latest has common cluster support for $\pi$ if
\begin{equation*}
    p(a|c, \pi_0) > 0\quad \forall a\in\A, c\in\C\mkern5mu:\mkern5mu p(a|c,\pi) > 0
\end{equation*}
Where we denote $$p(a|c,\pi) = \int_{\X} \pi(a|x)p(x|c) dx$$
\end{assumption}
\cref{as2} is weaker than \cref{as1} since for a given triplet $(x,c,a) \in \X \times \C \times \A$, the fact that $\pi_0(a|x) = 0, \pi(a|x) > 0$ does not ensure the same holds for every context within $c$. The idea of a homogeneous behaviour for every context inside a given cluster would make the CHIPS estimator circumvent the bias increase when Assumption \ref{as1} is not met for the IPS estimator (if \cref{as2} holds). Regarding the reward, this concept is formalized in the following assumption.

\begin{assumption}(\textit{Reward Homogeneity})
\label{as3}
We say that we observe reward homogeneity if the context $x$ does not affect on the reward $r$ given some action $a$ and some context $c$ (i.e., $r \bot x \mkern5mu|\mkern5mu c, a$).
\end{assumption}
The reward homogeneity assumption eliminates the dependency of the context on the reward when provided with the cluster and the action. Note that complying with Assumption \ref{as3} implies $q(a, c, x) = q(a, c, y) = q(a,c)$, where $x,y \in \X$, which together with Assumption \ref{as2} gives an alternative expression for the policy value in the following proposition:

\begin{proposition}
    Given a policy $\pi$, if Assumptions \ref{as2} and \ref{as3} hold, then we have that
    \begin{align}
        V(\pi) &:= \Exp{p(c)p(a|c, \pi)}{q(a,c)} \label{eq2}
    \end{align}
    Please refer to \cref{apdx:proof-p1} for a complete proof.
    \label{p1}
\end{proposition}

Considering the similarity of Equation \eqref{eq2} with the original policy value definition (Equation \eqref{policy-value}), Proposition \ref{p1} naturally motivates the analytical expression of the CHIPS estimator:
\begin{align*}
    \hat{V}_{\text{CHIPS}}(\pi; \mathcal{D}) &:= \frac{1}{N}\sum_{i=1}^N\frac{p(a_i|c_i, \pi)}{p(a_i|c_i, \pi_0)}r_i = \frac{1}{N}\sum_{i=1}^N w(a_i,c_i)r_i
\end{align*}

%%% -- reviewed till here --
\subsection{Theoretical Analysis}
First, we characterize the bias of the CHIPS estimator depending on the compliance with Assumptions \ref{as2} and \ref{as3}.
\begin{proposition}
    \label{p2}
    Under the Common Cluster Assumption (\ref{as2}) and the Cluster Homogeneity Assumption (\ref{as3}), the CHIPS estimator is unbiased for any given policy $\pi$:
    \begin{equation*}
        \Exp{\D}{\hat{V}_\text{CHIPS}(\pi; \D)} = V(\pi)
    \end{equation*} 
    Please refer to \cref{apdx:proof-p2} for a complete proof.
\end{proposition}

We note here that \cref{p2} implies that even when the Common Support Assumption (\ref{as1}) fails to ensure the unbiasedness of the IPS estimator, the CHIPS estimator can still use the more permissive Common Cluster Support (\ref{as2}), and the Reward Homogeneity (\ref{as3}) Assumption to ensure an unbiased estimate. Although \cref{as3} guarantees homogeneity at the reward level, a \emph{completely} homogeneous behaviour would also eliminate the context dependency at the action level, implying a deterministic policy given cluster, i.e. $p(a|c, \pi) = \pi(a|x) \mkern5mu \forall x \in c$. Both homogeneity conditions present a desirable scenario for the CHIPS estimator; however, they rarely occur when working in real-world data environments, which motivate the following assumption as a relaxation of the action-context independence:
\begin{assumption}(\textit{$\delta$-Homogeneity})
\label{as4}
Given a policy $\pi$, we say that the policy presents $\delta$-homogeneity if for any given action $a \in \A$, and any given cluster $c \in \C$, there exist $\delta^-_{\pi,c,a} \leq 1$ and $\delta^+_{\pi,c,a} \geq 1$ such that:
\begin{equation*}
    \delta^-_{\pi,c,a} \leq \frac{\pi(a | x)}{p(a | c, \pi)} \leq \delta^+_{\pi,c,a} \quad \forall x \in \X
\end{equation*}
\end{assumption}
It is worth noting that if $p(a | c, \pi) \neq 0 \mkern5mu \forall (x, c, a) \in \D$ then it is always possible to find $\delta^-_{\pi,c,a}$,  $\delta^+_{\pi,c,a}$ satisfying $\delta$-Homogeneity. The following proposition gives an upper bound for the bias of the CHIPS estimator when Assumption \ref{as3} cannot be ensured:

\begin{proposition}
    \label{p3}
    Given the logging data $\{(x_i, a_i, r_i)\}_{i=1}^N$ observed under some logging policy $\pi_0$, and an evaluation policy $\pi$ if the latest has common cluster support over the earliest, then we have that
    \begin{equation*}
    \left|\operatorname{Bias}\left(\hat{V}_{\text{\tiny{CHIPS}}}(\pi)\right)\right| \leq\left|\mathbb{E}_{p(c) p(x \mid c) p(a \mid c, \pi)}[q(a, c, x) \cdot \Delta_{c,a}]\right|
    \end{equation*}

    Where by \cref{as4} we have bounds ($\delta^-_{\pi,c,a}$,  $\delta^+_{\pi,c,a}$) for $\pi$,  ($\delta^-_{\pi_0,c,a}$,  $\delta^+_{\pi_0,c,a}$) for $\pi_0$, and we denote $\Delta_{c,a} = \max\{\delta^+_{\pi,c,a}, \delta^+_{\pi_0,c,a}\} - \min\{\delta^-_{\pi,c,a}, \delta^-_{\pi_0,c,a}\}$. Please refer to \cref{apdx:proof-p3} for a complete proof.
\end{proposition}

Proposition \ref{p3} formalizes the intuition on how the bias of the estimator under Assumption \ref{as2} depends on the extent to which the contexts inside a cluster behave homogeneously under a given policy. Formally, the gap $\delta^+_{\pi} - \delta^-_{\pi}$ determines how close the CHIPS is to being unbiased, being the case $\delta^-_{\pi,c,a} = \delta^+_{\pi,c,a} = 1$ the perfect scenario. In this case, we have that $\pi(a | x) = p(a | c, \pi)$, which means that the weights in IPS $w(a,x) = w(a,c)$, and we could in theory substitute any context for any other within the same cluster for calculations, mitigating the problems that arise when Assumption \ref{as1} does not hold. Additionally, we can also provide an expression for the difference in mean squared error with respect to IPS in the same conditions as \cref{p3}:
\begin{proposition}
    \label{p8}
    Under the same conditions as in \cref{p3}, the difference in mean squared error between CHIPS and MIPS can be expressed as
    \begin{align*}
        \operatorname{MSE}\left(\hat{V}_{\text{IPS}}(\pi)\right) 
        & - \operatorname{MSE}\left(\hat{V}_{\text{CHIPS}}\right) \\
        & = \mathbb{V}_D\left[\hat{V}_{\text{IPS}}(\pi)\right] 
        - \mathbb{V}_D\left[V_{\text{CHIPS}}(\pi ; D)\right] \\
        & \quad - \operatorname{Bias}\left(\hat{V}_{\text{CHIPS}}(\pi)\right)^2
    \end{align*}
    
    Please refer to \cref{apdx:proof-p8} for a complete proof.
\end{proposition}

It is also worth studying the bias of the CHIPS estimator when the Common Cluster Support assumption does not hold, while the Assumption \ref{as3} holds. For this purpose, we acknowledge that the bias of the IPS estimator when Assumption \ref{as1} is not met can be given in terms of the actions violating such assumption \citep{Sachdeva2020}:
\begin{equation*}
    \left\lvert\text{Bias}(\hat{V}_{\text{IPS}}(\pi;\D))\right\rvert = \Exp{p(x)}{\sum_{\mathcal{U}(x, \pi_0)} \pi(a|x)q(a,c,x)}
\end{equation*}

Where $\mathcal{U}(x, \pi_0) := \{a \in \A\mkern5mu|\mkern5mu\pi_0(a, x) = 0\}$ are known as the \textit{deficient} actions. Following a similar approach we introduce the following proposition:

\begin{proposition}
    \label{p4}
     Given the logging policy $\pi_0$ and some evaluation policy $\pi$, the absolute bias of the CHIPS estimator when Assumption \ref{as3} holds can be expressed as
    \begin{equation*}
        \left\lvert\text{Bias}(\hat{V}_{\text{CHIPS}}(\pi;\D))\right\rvert = \Exp{p(c)}{\sum_{\mathcal{U}(c, \pi_0)} p(a| \pi, c)q(a, c)}
    \end{equation*}
    Where $\mathcal{U}(c, \pi_0) := \{a \in \A\mkern5mu|\mkern5mup(a|\pi_0, c) = 0\}$. Please refer to \cref{apdx:proof-p4} for a complete proof.
\end{proposition}

\begin{corollary}
\label{cor:1}
Under the conditions of Proposition \ref{p4}, we have that
\begin{align*}
    &\left\lvert\text{Bias}(\hat{V}_{\text{IPS}}(\pi;\mathcal{D}))\right\rvert 
     - \left\lvert\text{Bias}(\hat{V}_{\text{CHIPS}}(\pi;\mathcal{D}))\right\rvert \\
    & \phantom{=}=\mathbb{E}_{p(c)} \left[ \sum_{\mathcal{U}(x, \pi_0) \setminus \mathcal{U}(c, \pi_0)} p(a \mid \pi, c) q(a, c) \right].
\end{align*}

Please refer to Appendix \ref{proof:p4} for a complete proof.
\end{corollary}

Note that in this case, the CHIPS' reduction in absolute bias depends directly on the number of actions that violate \cref{as1}, but still comply with Assumption \ref{as3}. Thus, the greater the number of deficient actions by Common Support condition covered by the Common Cluster Support, the more significant the bias reduction with respect to IPS.
In this conditions, its also interesting to study the difference in bias with respect to the other two transformation-based methods (MR and MIPS), a result given by the next proposition:
\begin{proposition}
    \label{p9}
    Let $f_{\phi^*}$ be defined as in \cref{eq:mr} with $f_{\phi^*} = w(a,x) + \epsilon$ for some $\epsilon \in \mathbb{R}$ and $e\in\mathcal{E}$ give action embeddings. Under the conditions of the \cref{p4}, we have that:
\begin{align*}
    & \left|\operatorname{Bias}\left(\hat{V}_{\text{MR}}; \mathcal{D}\right)\right| - \left|\operatorname{Bias}\left(\hat{V}_{\text{CHIPS}}; \mathcal{D}\right)\right| \\
    & = -\mathbb{E}_{p(c)}\left[\sum_{a \in \left(\mathcal{U}\left(c, \pi_0\right) \backslash \mathcal{U}\left(c, \pi_0\right)\right)} q(a, c) p(a \mid \pi, c)\right] \\
    & \quad + \epsilon \mathbb{E}_{p(c)}\left[\sum_{a \in \mathcal{A}} q(a, c) p\left(a \mid \pi_0, c\right)\right] \\[1em]
    & \left|\operatorname{Bias}\left(\hat{V}_{\text{MIPS}}; \mathcal{D}\right)\right| - \left|\operatorname{Bias}\left(\hat{V}_{\text{CHIPS}}; \mathcal{D}\right)\right| \\
    & = \mathbb{E}_{p(x)}\left[\sum_{e \in \mathcal{U}\left(e, \pi_0\right)} p(e \mid x, \pi) q(x, e)\right] \\
    & \quad - \mathbb{E}_{p(c)}\left[\sum_{a \in \mathcal{U}\left(c, \pi_0\right)} p(a \mid c, \pi) q(a, c)\right].
\end{align*}

    Please refer to \cref{apdx:proof-p9} for a complete proof.
\end{proposition}
When studying homogeneity at an action level, we have focused on the probability of observing an action for a particular context $x$ within a cluster $c$ (i.e., $\pi(a|x)$). Conversely, we can also study the \textit{predictability} of a context given an action and a cluster under a policy $\pi$, which we denote as $p(x|a,c) = \pi(x|a, c)$. Ideally, we would have that the conditional probability distribution of the context given the action and the cluster is uniform (i.e., $\pi(x_i|a, c) = \pi(x_j|a, c) \mkern5mu\forall\mkern5mu x_i,x_j \in c$). Predictability is used in the following proposition, that characterizes the relation between the reduction in variance of the CHIPS estimator with respect to IPS:

\begin{proposition}
    \label{p6}
    Given a logging policy $\pi_0$, under the Common Support Assumption (\ref{as1}) and the Reward Homogeneity Assumption (\ref{as3}) we have that
    \begin{align*}
    & N\left(\Var{\D}{\hat{V}_{\text{IPS}}(\pi;\D)} - \Var{\D}{\hat{V}_{\text{CHIPS}}(\pi;\D)}\right) \\
    & = \Exp{p(c)p(a|c,\pi_0)}{\Var{\pi_0(x|a, c)}{w^2(a,x)} \Exp{p(r|a, c)}{r^2}}.
    \end{align*}

    Note that this quantity is always positive, implying that CHIPS always reduces the variance of IPS.
    Please refer to \cref{apdx:proof-p6} for a complete proof.
\end{proposition}

Proposition \ref{p6} indicates that when Assumptions \ref{as1} and \ref{as3} hold, the variance reduction of CHIPS compared to IPS corresponds to the total decrease in mean squared error when approximating the actual policy value $V(\pi)$, as both estimators are unbiased under these conditions. This mean squared error gap is influenced by two factors: First, $\Exp{p(r|a, c)}{r^2}$, reflecting the noise in rewards for actions within the same cluster (related to Assumption \ref{as3}). Second, the variance of IPS weights conditioned on the predictability $p(x|a,c)$, which increases when $w(a,x)$ varies widely (e.g., when logging and evaluation policies differ) or when $\pi(x|a,c)$ is uninformative (contexts behave uniformly given the cluster and action). Thus, the variance reduction in CHIPS is particularly pronounced when IPS exhibits high variance and contexts within a cluster are similar.
Furthermore, if MIPS and CHIPS are in the same space (considering contexts $c\in\mathcal{C}$ as described and action embeddings $e\in\mathcal{E}$), \cref{p6} can be extended to show that CHIPS has less variance than MIPS:

\begin{proposition}
    \label{p7}
    In context-action-embedding joint space ($\mathcal{X}\rightarrow\mathcal{C}\rightarrow\mathcal{A}\rightarrow\mathcal{E}\rightarrow[0, R_{max}]$), if Assumptions \ref{as2} and \ref{as3} hold, as well as their MIPS counterparts (\textit{Common Embedding Support} and \textit{No Direct Effect}), then we have that
    \begin{equation*}
        \mathbb{V}_{\D}(\hat{V}_{\text{IPS}}(\pi)) \geq
        \mathbb{V}_{\D}(\hat{V}_{\text{MIPS}}(\pi)) \geq
        \mathbb{V}_{\D}(\hat{V}_{\text{CHIPS}}(\pi)) \geq 0
    \end{equation*}
    Please refer to \cref{apdx:proof-p7} for a complete proof.
\end{proposition}

\subsection{Empirical Calculations}
\label{subs:3.2}
The alternative analytical expression for the policy value given in Equation \ref{eq2} eliminates the dependency on the original definition of policy value and motivates the CHIPS estimator under assumptions \ref{as2} and \ref{as3}. However, in practice, assessing if such conditions hold is complicated, particularly if we have limited logging data. To mitigate this problem and justify using CHIPS in real-world settings, we need to make an approximation to context-homogeneous behavior on both action and reward levels within a cluster. In practice, we have a clustering method $\xi: \X \rightarrow \C$, and we use the transformation:
\begin{align*}
  \tau \colon (\X, \A, [0, R_{max}]) &\to (\X, \C, \A, [0, R_{max}])\\\
  (x,a,r) &\mapsto (x, \xi(x), a, r).
\end{align*}
Given a policy $\pi$ and a cluster $c$, we use the definition to estimate $p(a | c, \pi)$:
\begin{align}
    p(a|c,\pi) &= \int_{\X} \pi(a|x)p(x|c) \, dx \nonumber \\
    &= \int_{x \in c} \pi(a|x)p(x|c) \, dx \nonumber \\
    &\approx \frac{1}{\left|\D_c\right|}\sum_{x \in \D_c} \pi(a|x), \label{eq3}
\end{align}
Here, we denote $\D_c = \{(x,\tilde{c},a,r) \in \tau(\D) \mkern5mu:\mkern5mu \tilde{c} = c\}$. In Equation \ref{eq3}, we used that $p(x|c) = 0$ if $c \neq \xi(x)$. Since this equation is essentially $\Exp{p(x|c)}{\pi(a|x)}$, we approximate this value by averaging $\pi(a|x)$ over all contexts inside the given cluster.

The second approximation needed involves the reward being independent of the context given the action and the cluster, i.e., $q(a,c,x) = q(a,c)$. Following a similar approach than in the previous case, for a particular (given) action $a$ and cluster $c$, we observe that $q(a,c) = \Exp{p(x|c)}{\pi(r|a, c, x)}$, which motivates the idea of an \textit{average reward} per cluster. In our synthetic experiments, the reward is binary, therefore we will assume that the observations inside a cluster are observations in a Bernoulli process (i.e., $R_c \sim \text{Ber}(\theta)$) and estimate this average reward using two different approaches:
\begin{itemize}[leftmargin=*]
    \item \textbf{Maximum Likelihood (ML)} In which we just average the rewards observed within a cluster $c$ for each action $a$ as $\hat{r}_{\text{mean}}(a,c) = \frac{1}{\left|R_c\right|}\sum_{R_c} r_k$ with $R_c:=\{r_k : (x_k,c_k,a_k,r_k) \in \D_c\}$.
    \item \textbf{Maximum A Posteriori (MAP)}. In this setting, estimating the average reward is equivalent to estimating the most probable $\theta$ using a beta prior, where we obtain:
    \begin{equation*}
         \hat{r}_{\text{bayes}}(\alpha, \hat{\beta}; c) = \frac{(\alpha - 1) + \sum_{R_c}r_k}{\alpha + \hat{\beta} + |R_c| - 2}
    \end{equation*}
    Where we denote $\alpha, \hat{\beta}$ as the parameters of the prior Beta distribution. In our experiments, we use non-informative priors ($\alpha = \hat{\beta}$) \citep{comparison, kerman} and we explore the choosing of this parameter for arbitrary problems in \cref{apdx:alphas}. Please refer to \cref{apdx:reward-estimates} for the complete derivations of the MAP and ML estimations.
\end{itemize}
\section{EXPERIMENTS}
\label{section:4}
\subsection{Synthetic dataset}
We compare CHIPS with other baseline estimators (IPS, DM, DR, SNIPS \citep{snips}, DRoS \citep{dros}, SNDR \citep{sndr}, MR \citep{taufiq2023marginal}) in estimating the evaluation policy value in a cluster-based synthetic dataset in which we can control the difficulty of the OPE problem. A description of all hyperparameters used for generation (e.g., $a_{num}$, $c_{exp}$ ...) can be found in \cref{apdx:params}. We start by generating cluster centers $\C := \{c_k\}_{k=1}^m$ inside a $d_x$-dimensional 
% random 
ball $B(0,\text{c}_{\text{exp}}):=\{x \in \R^{d_x}\mkern5mu:\mkern5mu{||x||}^2 < \text{c}_{\text{exp}}\}$ using a variation of the Box-Muller transformation \citep{box-muller}:
\begin{equation*}
    c_k = \frac{c_{\text{exp}} \cdot {u_k}^{-dx} \cdot z_k}{||z_k||},
\end{equation*}
where $U:=\{u_k\}_{k=1}^m\sim U[0,1]$ and $Z:=\{z_k\}_{k=1}^m \sim \mathcal{N}(0, \mathbb{1}_{d_x})$. 
We sample $S:=\{s_k\}_{k=1}^m\sim U[0,1]$, and use the softmax transformation $\phi(\mathcal{S})$ to define $p(c_i) = \phi(\mathcal{S})_i$. 
Then, we sample cluster centers according to this distribution $w=\{w_i\}_{i=1}^{x_\text{num}} \sim \phi(\mathcal{S})$, and, for each center $c_i$, we uniformly sample points belonging to the $n$-ball centered on $c_i$, using the same variation of the Box-Muller transform that we used previously:
\begin{equation*}
\X_i = (x_i^1,..., x_i^{h_i}) \sim U[B(c_i, c_{\text{rad}})]
\end{equation*}
Note here that $h_i = \sum_{i=1}^{x_\text{num}} \mathbb{1}_{\{c_i=w_i\}}$. 
We define the context space as the union of these generated points $\X = \bigcup_{i=1}^{m}{\X_i}
=\{x_i\}_{i=1}^{x_\text{num}}$. We sample $\mathcal{V}=\{v_i\}_{i=1}^{x_\text{num}} \sim \mathcal{N}(0,1)$ and define $p(x_i) = \phi(\mathcal{V})_i$ using the $\phi$ softmax transformation again. We then use these probabilities to sample the logging ($\X_{\text{log}}$) and evaluation ($\X_{\text{eval}}$) data, with $|\X_{\text{eval}}| = e_{len}$ and $|\X_{\text{log}}| = b_{len}$. To generate the policies, we sample  $y_i=\{y_i^j\}_{j=1}^{a_\text{num}} \sim \mathcal{N}(0,1)$ for every cluster $c_i$ 
(where $a_\text{num}$ is the number of actions) and $z=\{z_k\}_{i=1}^{x_\text{num}} \sim \mathcal{N}(0,1)$ to define the policies for every context in cluster $c_i$ as:
\begin{align*}
    \pi(a_j | c_i, x_k) &= \frac{e^{y_i^j + \sigma z_k}}{\sum_{m=1}^{a_\text{num}} e^{y_i^m + \sigma z_k}} \\
    \pi_0(a_j | c_i, x_k) &= \frac{e^{\beta(y_i^j + \sigma z_k)}}{\sum_{m=1}^{a_\text{num}} e^{\beta(y_i^m + \sigma z_k)}}, \quad -1 \leq \beta \leq 1
\end{align*}
Given a context $x_k$, both policies are determined by a term that depends on the cluster and the action ($u_i^j$), and a term that depends on the context itself ($x_k$). Here $0 \leq \sigma \leq 1$ controls how independent a policy is from the context and $\beta$ how close the logging and evaluation policies are. For obtaining the actions, we sample $\A_{\text{log}} \sim \pi_0$ and $\A_{\text{eval}} \sim \pi$. For generating the rewards, we create a misspecified reward setting by defining:
\begin{equation*}
    r(a_i,c_i,x_i) = \mathbb{1}\left\{u_i < \pi(a_i|c_i,x_i) \cdot \frac{||x_i||_1}{c_{\text{exp}}d_x}\right\}, 
\end{equation*}
where $u_i\sim U[0,1]$.
The reward depends on two factors; the first one is the Manhattan norm of the context; the further from 0, the more
likely it is to observe a positive reward. The second factor is the evaluation policy $\pi(a_i|c_i,x_i)$, which makes
this a misspecified reward setting when the logging and evaluation policies are different enough. In this case, the $(a_i,c_i,x_i)$ triplets having the highest probability of observation under the evaluation policy are more likely to observe positive rewards, resulting in a significant difference with respect to the observed rewards under the logging policy for such triplets. We sample rewards using this method for the logging ($\mathcal{R}_{\text{log}}$) and evaluation ($\mathcal{R}_{\text{eval}}$) data to obtain $\D_{\text{log}}:=(\X, \C, \A_{\text{log}}, \mathcal{R}_{\text{log}})$ and $\D_{\text{eval}}:=(\X, \C, \A_{\text{eval}}, \mathcal{R}_{\text{eval}})$. Finally, we select a subset for $N$ samples from both sets. A representation of the generated structure can be found in \cref{fig:synthetic-dataset}.
%% Reviwed till here %%
\subsubsection{Synthetic results}
\label{sec:synthetic results}
In this section we analyze CHIPS performance while varying parameters of the synthetic dataset. In our experiments, the generation process for each parameter value is repeated 100 times with different random seeds. The final reported results are the average over all experiments, with the standard deviation corresponding to the lighter bands represented in all the figures. The basic configuration for the parameters used throughout the experiments can be found in \cref{apdx:params}, along with the specifications of the hardware used. We use Random Forest \citep{random-forest} to obtain $\hat{q}(x,a)$ in DM-based methods and mini-batch KMeans \citep{kmeans} implementation in SciKit-Learn \citep{scikit-learn} as the clustering method for CHIPS (alternative clustering methods and their performance are also discussed in \cref{apdx:extra}). We also use $\beta = -1$, maximizing the distributional shift between logging and evaluation policies.

\begin{figure*}[ht]
    \centering
    \includegraphics[width=\textwidth]{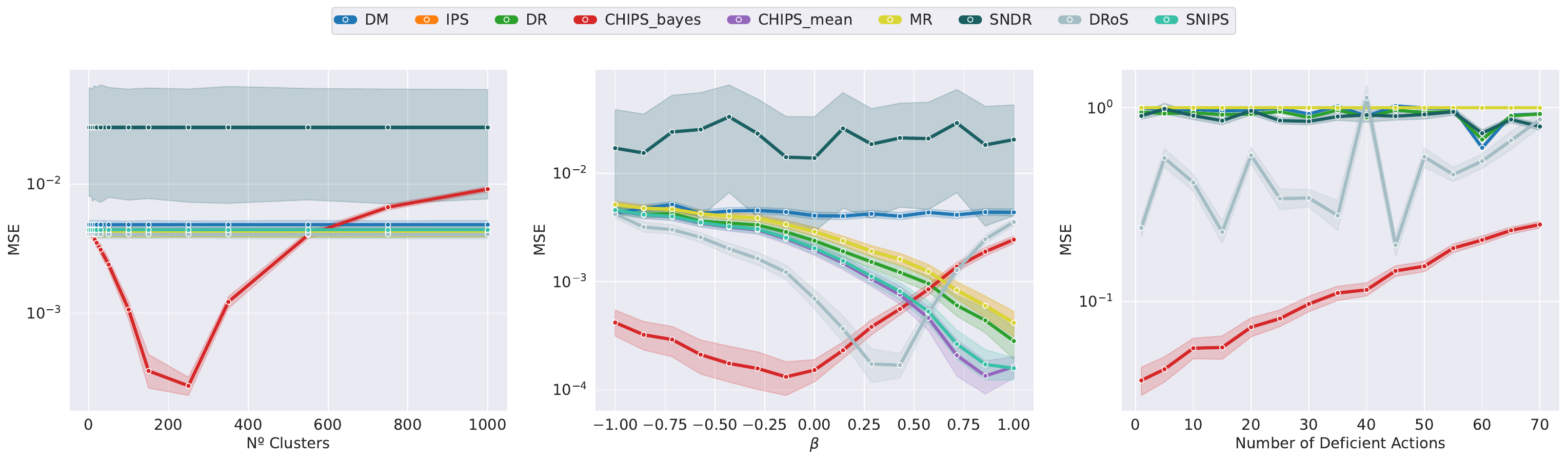}
    \caption{From left to right, the mean square error in the synthetic dataset experiments varying the number of clusters, the distributional shift between logging and evaluation policy ($\beta$), and the number of deficient actions in the logging data (normalized w.r.t. IPS).}
    \label{fig:syn}
\end{figure*}

\textbf{Number of clusters.} For this experiment, we vary the number of clusters the CHIPS estimator uses, with values ranging from 1 to 1000. Since $\beta = -1$, the implementation of CHIPS using ML reward estimation is unsuccessful (see \cref{apdx:map-v-ml} for a further discussion). On the other hand, for the MAP case, we observe a v-shaped error graph (see \cref{fig:syn} (left)), suggesting that CHIPS performance is sensitive to effectiveness of clustering. In particular, we have a highly biased estimation when assuming insufficient or excessive clusters (see \cref{fig:clusters}). The reason for this bias in the first case might be an oversimplification of the structure of the cluster space. Conversely, we progressively gain bias when we select too many clusters according to Proposition \ref{p4} as CHIPS converges to IPS. In this case, CHIPS is also vulnerable to reward misspecification, which causes an increase in variance. In practice, this parameter can be selected by considering the possible CHIPS estimates as a parametric family depending on the number of clusters and use the PAS-IF technique \citep{pasif} to choose the optimal number of clusters.

\textbf{Beta.} This experiment examines the impact of the distributional policy shift between $\pi$ and $\pi_0$. Lower values in our range (i.e., $\pi_0 \longleftrightarrow \pi$) result in significant policy shifts that introduce bias in IPS estimates for large context-action spaces \citep{mips, Sachdeva2020}. The CHIPS estimator mitigates this by treating all context-action samples within a cluster as if they share the same context. However, when $\beta$ is low, these virtual extra samples may not suffice for accurate estimation, as the most relevant $(x,a)$ pairs ($\pi(a|x)$ near 1) are underrepresented (see \cref{apdx:map-v-ml}). In such cases, ML estimation in CHIPS is ineffective, while MAP estimation provides some resistance by pushing reward estimates towards the posterior expectation, making it sensitive to prior choice. However, this resistance can be counterproductive when the distributional shift is small ($\beta$ close to 1), as both ML estimates and IPS converge faster to more accurate estimations (see \cref{fig:syn} (center)).

\textbf{Deficient actions.} In this setting we explicitly set the probability ($\pi_0$) of observing a variable number of actions in the action space to 0 and evaluate CHIPS' response in a space with 200 actions and $\beta=-1$. This setting is quite challenging as not only we have deficient actions but also a significant distributional shift between policies. The majority of baselines perform at a similar level than IPS with the exception of DRoS \citep{dros}, that performs slightly better but is still outperformed by CHIPS.

Additional experiments and discussions of results varying other parameters, different clustering methods, and a time complexity analysis can be found in \cref{appdx:time}.

\subsection{Real dataset}
Following the literature, for assessing the capabilities of the CHIPS estimator in a real-world environment, we compare the performance in the Open Bandit Dataset (OBD) \citep{obd} of IPS, DM, DR, MRDR \citep{mrdr} and MIPS \citep{mips}, with and without SLOPE \citep{slope}. The OBD dataset was gathered using two different policies during an A/B test: uniform random, which we consider as logging (i.e., $\pi_0$), and Thompson sampling \citep{Thompson1933, Thompson1935}, which we consider as evaluation (i.e., $\pi$). The dataset is based on a recommendation system for fashion e-commerce. We observe user data as contexts $x$, items to recommend $a\in\A$ (with $|\A|=240$) and rewards $r \in \{0, 1\}$ representing user interactions. 

\begin{figure*}[ht]
    \centering
    \includegraphics[width=\textwidth]{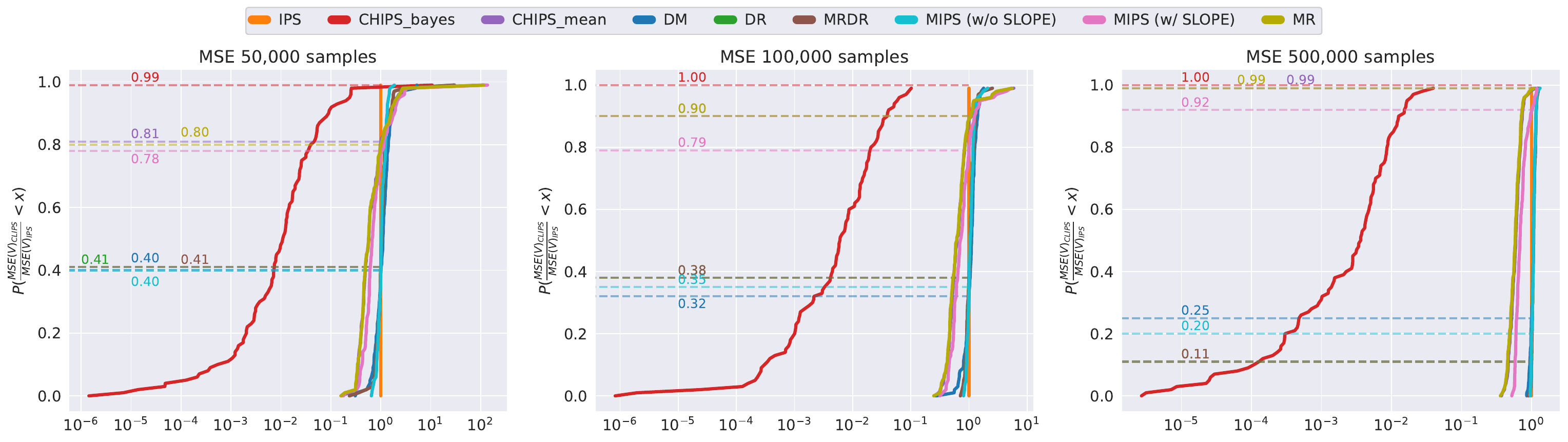}
    \caption{ECDF of the relative mean squared error with respect to IPS for the real dataset using 50000 (left), 100000 (center), and 500000 (right) logging samples.}
    \label{fig:real}
\end{figure*}

Following the experimental protocol of \citet{mips} (see \cref{apdx:experimental-protocol}), we experiment with the real dataset varying the number of logging samples available for the estimation using $50\,000$, $100\,000$, and $500\,000$ samples to compute the Empirical Cumulative Distribution Function (ECDF) of the normalized mean squared error with respect to IPS. We increase the number of clusters for CHIPS as more logging samples are available to try to maximize performance, following the intuition from our earlier experiments on the synthetic dataset (see \cref{fig:bivariate} (right)). We use 8 clusters for $100\,000$ samples as a reference from our results for 240 actions in the synthetic dataset (see \cref{fig:bivariate} (left)). Regarding the clustering method, we use again mini-batch KMeans. 

We observe that the CHIPS estimator using the ML approximation is slightly better (+3\%) than MIPS when few samples are available (see \cref{fig:real}, (left)). This performance gap widens (+11\%) as the CHIPS estimator has more samples available (see \cref{fig:real}, (center)) and starts narrowing (+7\%) as the number of samples is enough for MIPS to also start making more accurate estimations (see \cref{fig:real}, (right)). 

Using the MAP reward estimation for CHIPS provides a considerable advantage in all experiments since the real dataset present severe reward misspecification, as discussed in \cref{apdx:map-v-ml}. Similarly to the synthetic dataset, the partition structure of the cluster space and the $\alpha$ parameter in MAP are sensitive parameters. In particular, for the number of clusters, we observe that using an insufficient or excessive number of clusters can negatively impact performance (see \cref{fig:real-extra} (left)) as we discussed in section \cref{sec:synthetic results}. Regarding the value of $\alpha$ for the Beta prior, following the results from the synthetic experiment studying the effect of this parameter conjointly with the distributional shift between logging and evaluation policies (see discussion in \cref{apdx:alphas} and \cref{fig:alpha_beta}), we used $\alpha=20$ as the logging policy is uniform (the equivalent of $\beta = 0$ in the synthetic dataset). \cref{fig:real-extra} (right) shows how choosing a lower or higher value for $\alpha$ deteriorates the performance of the CHIPS estimator, reaffirming the results observed in the synthetic dataset (see \cref{fig:alpha_beta}).

\section{CONCLUSIONS, LIMITATIONS AND FUTURE WORK}
\label{section:5}
In this work we have explored an alternative approach to the OPE problem by clustering contexts instead of pooling information over actions to mitigate the problems arising in 
IPS when the Common Support condition does not hold. The proposed setup for the OPE problem using contexts led to the CHIPS estimator, which uses a similar approach to IPS applied over clusters instead of contexts. We have studied this estimator extensively from a theoretical and practical perspective, evaluating its performance for different configurations in a controlled synthetic dataset and a real-world example. The results obtained in the experiments for both cases demonstrate that the CHIPS estimator provides a significant improvement in estimation accuracy, outperforming existing estimators if the context space has a cluster structure. The accuracy of CHIPS is also influenced by the accuracy of the clustering method and the homogeneity behaviour of contexts inside the same cluster. Additionally, choosing a balanced number of clusters to avoid over- and under-simplification of the cluster structure is an important part of the estimation process and opens the possibility of exploring if it is possible to estimate the optimal value for hyperparameters beyond empirical estimation or even if combining CHIPS with pure action-embedding methods like MIPS can improve general performance.

%\subsubsection*{Acknowledgements}
%All acknowledgments go at the end of the paper, including thanks to reviewers who gave useful comments, to colleagues who contributed to the ideas, and to funding agencies and corporate sponsors that provided financial support. 
%To preserve the anonymity, please include acknowledgments \emph{only} in the camera-ready papers.

\bibliographystyle{aaai25}
\bibliography{aistats25_2100_clustering_contexts_in_ope}

%%%%%%%%%%%%%%%%%%%%%%%%%%%%%%%%%%%%%%%%%%%%%%%%%%%%%%%%%%%%
\section*{Checklist}

 \begin{enumerate}

 \item For all models and algorithms presented, check if you include:
 \begin{enumerate}
   \item A clear description of the mathematical setting, assumptions, algorithm, and/or model. 
   
   [\textbf{Yes}]. \textbf{Justification:} The mathematical setting of both synthetic and real experiments is detailed in \cref{section:4}, the theoretical assumptions, its analysis when they don't hold, and the main algorithm are detailed in \cref{section:3}.

   \item An analysis of the properties and complexity (time, space, sample size) of any algorithm.

    [\textbf{Yes}]. \textbf{Justification:} A complete analysis on the complexity of the method can be found in \cref{appdx:time}, while the analysis of properties can be found in \cref{section:3}.

   \item (Optional) Anonymized source code, with specification of all dependencies, including external libraries. [Yes/No/Not Applicable]

   [\textbf{Yes}]. \textbf{Justification:} The code, an explanation on how to execute it and a Poetry environment to take care of the dependencies are included in the supplemental materials.

 \end{enumerate}

 \item For any theoretical claim, check if you include:
 \begin{enumerate}
   \item Statements of the full set of assumptions of all theoretical results.
   
   [\textbf{Yes}]. \textbf{Justification:} The full set of assumptions, relaxation of them and all the theoretical analysis of the method can be found in \cref{section:3}.
   
   \item Complete proofs of all theoretical results.

    [\textbf{Yes}]. \textbf{Justification:} For every proposition in the theoretical analysis (\cref{section:3}), the proof is referenced and can be found in \cref{apdx:theoretical-proofs}.
    
   \item Clear explanations of any assumptions.

   [\textbf{Yes}]. \textbf{Justification:} For every proposition in the theoretical analysis (\cref{section:3}), the assumptions used for the results are detailed in the introduction of the proposition.
   
 \end{enumerate}

 \item For all figures and tables that present empirical results, check if you include:
 \begin{enumerate}
   \item The code, data, and instructions needed to reproduce the main experimental results (either in the supplemental material or as a URL). 

  [\textbf{Yes}]. \textbf{Justification:} The code contains the necessary scripts to reproduce every experiment in the paper, as well as instructions on how to execute it.
   
   \item All the training details (e.g., data splits, hyperparameters, how they were chosen). 

   [\textbf{Yes}]. \textbf{Justification:} The training details and parameter variations can be found in \cref{section:4}. Extra experiments varying more parameters, 2 parameters at the same time or different clustering options can be found in \cref{apdx:extra}. Additionally, a table with the base value of every parameter can be found in \cref{apdx:params}.
   
    \item A clear definition of the specific measure or statistics and error bars (e.g., with respect to the random seed after running experiments multiple times).
    
    [\textbf{Yes}]. \textbf{Justification:} All deviation bars obtained after 100 runs with different seeds for every parameter are included in the result figures as explained in \cref{section:4}.
    
     \item A description of the computing infrastructure used. (e.g., type of GPUs, internal cluster, or cloud provider).

    [\textbf{Yes}]. \textbf{Justification:} The table with the specifications of the computing infrastructure used for the experiments can be found in \cref{tab:resources}.
     
 \end{enumerate}

 \item If you are using existing assets (e.g., code, data, models) or curating/releasing new assets, check if you include:
 \begin{enumerate}
   \item Citations of the creator If your work uses existing assets. 
    
   [\textbf{Yes}]. \textbf{Justification:} The evaluation pipeline that we use for the real experiments, the systems that we compare our method with, and the code implementations of known algorithms for different libraries that we use are referenced in the paper.
   
   \item The license information of the assets, if applicable.
   
   [\textbf{Yes}]. \textbf{Justification:} The license included in the code does not conflict with the license of the used resources.
   
   \item New assets either in the supplemental material or as a URL, if applicable. 
   
   [\textbf{Yes}]. \textbf{Justification:} We provide all the code with our pipeline and implementations for common ope algorithms in the supplemental material.
   
   \item Information about consent from data providers/curators.
   
   [\textbf{Yes}]. \textbf{Justification:} The code includes a licence to use.
   
   \item Discussion of sensible content if applicable, e.g., personally identifiable information or offensive content.
    
    [\textbf{Not Applicable}]. \textbf{Justification:} There is not personally identifiable information or offensive content of any kind in our data or experiments.
 \end{enumerate}

 \item If you used crowdsourcing or conducted research with human subjects, check if you include:
 \begin{enumerate}
   \item The full text of instructions given to participants and screenshots. 
   
   [\textbf{Not Applicable}]. \textbf{Justification:} Our research did not use crowdsourcing or was conducted with human subjects.
   
   \item Descriptions of potential participant risks, with links to Institutional Review Board (IRB) approvals if applicable.
   
   [\textbf{Not Applicable}]. \textbf{Justification:} Our research did not use crowdsourcing or was conducted with human subjects.
   
   \item The estimated hourly wage paid to participants and the total amount spent on participant compensation.
   
  [\textbf{Not Applicable}]. \textbf{Justification:} Our research did not use crowdsourcing or was conducted with human subjects.
 \end{enumerate}

 \end{enumerate}

\appendix
\onecolumn
\section{THEORETICAL RESULTS PROOFS}
\label{apdx:theoretical-proofs}
\subsection{Proposition \ref{p1}}
\label{apdx:proof-p1}
Given a policy $\pi$, if both Assumption \ref{as2} and \ref{as3} hold, from the refinement of the policy value definition in a cluster-based bandits process (introduced in Section \ref{section:3}), we have that:
\begin{align}
     V(\pi) &:= \Exp{p(x)p(c|x)\pi(a|x)p(r|a, c, x)}{r} \nonumber\\
            &= \Exp{p(c)p(x|c)\pi(a|x)}{q(a,c,x)} \label{eq:5} \\
            &= \Exp{p(c)}{\int_{\X}p(x|c)\sum_{a \in \A}p(a|c)q(a,c) dx} \label{eq:6}\\
            &= \Exp{p(c)}{\int_{\X}\sum_{a \in \A}p(x|c)\pi(a|x)q(a,c) dx} \nonumber\\
            &= \Exp{p(c)}{\sum_{a \in \A}q(a,c)\int_{\X}p(x|c)\pi(a|x) dx} \nonumber\\
            &= \Exp{p(c)}{\sum_{a \in \A}p(a|c,\pi)q(a,c)} \label{eq:7}\\
            &= \Exp{p(c)p(a|c,\pi)}{q(a,c)}\nonumber
\end{align}
Where in Equation \ref{eq:5} we used the Bayes Theorem, in Equation \ref{eq:6} the fact that under Assumption \ref{as3} $q(a,c,x) = q(a,c)$, and the definition of $p(a|c,\pi)$ in Equation \ref{eq:7}.

\subsection{Proposition \ref{p2}}
\label{apdx:proof-p2}
Given a policy $\pi$ and under Assumptions \ref{as2} and  \ref{as3} we have that:
\begin{align}
     \Exp{\D}{\hat{V}_\text{CHIPS}(\pi; \D)} &= \Exp{\D}{w(a,c)r} \label{eq:8}\\
    &=\Exp{p(x)p(c|x)\pi_0(a|x)p(r|a,c,x)}{w(a,c)r} \nonumber\\
    &=\Exp{p(c)p(x|c)\pi_0(a|x)}{w(a,c)q(a,c)} \label{eq:9}\\
    &=\Exp{p(c)}{\int_{\X}p(x|c)\sum_{a\in\A}\pi_0(a|x)w(a,c)q(a,c)dx}\nonumber\\
    &=\Exp{p(c)}{\int_{\X}\sum_{a\in\A}p(x|c)\pi_0(a|x)w(a,c)q(a,c)dx}\nonumber\\
    &=\Exp{p(c)}{\sum_{a\in\A}w(a,c)q(a,c)\left(\int_{\X}p(x|c)\pi_0(a|x)dx\right)}\nonumber\\
    &=\Exp{p(c)}{\sum_{a\in\A}\frac{p(a|c,\pi)}{\cancel{p(a|c,\pi_0)}}q(a,c)\cancel{p(a|c,\pi_0)}}\label{eq:10}\\
    &=\Exp{p(c)p(a|c,\pi)}{q(a,c)}\nonumber\\
    &=\Exp{p(x)p(c|x)\pi(a|x)p(r|a,c,x)}{r}\label{eq:11}\\
    &=V(\pi)\nonumber
\end{align}
In Equation \ref{eq:8}, we have used the linearity of expectation, in Equation \ref{eq:9} the definition of $q(a,c,x)$ and Assumption \ref{as3}. Equation \ref{eq:10} is just using the definition of $p(a|c,\pi)$ while Equation \ref{eq:11} is a combination of Proposition \ref{p1} and the equivalence $q(a,c)=q(a,c,x)$ under the given assumptions.

\subsection{Proposition \ref{p3}}
\label{apdx:proof-p3}
Given the logging data $\D=\{(x_i,a_i,r_i)\}$, a logging policy $\pi_0$, and an evaluation policy $\pi$ having common cluster support over it, we have that:
\begin{align}
    \text{Bias}(\hat{V}_{\text{CHIPS}}(V; \D)) &= \Exp{\D}{w(c,a)r} - V(\pi)\nonumber\\
    &=\Exp{p(x)p(c|x)\pi_0(a|x)p(r|a,c,x)}{w(a,c)r} - V(\pi)\nonumber\\
    &=\Exp{p(x)p(c|x)\pi_0(a|x)}{w(a,c)q(a,c,x)} - V(\pi)\nonumber\\
    &=\Exp{p(x)p(c|x)\pi_0(a|x)}{w(a,c)q(a,c,x)} - \Exp{p(x)p(c|x)\pi(a|x)}{q(a,c,x)}\nonumber\\
    &=\Exp{p(c)p(x|c)\pi_0(a|x)}{w(a,c)q(a,c,x)} - \Exp{p(c)p(x|c)\pi(a|x)}{q(a,c,x)}\nonumber\\
    &=\Exp{p(c)}{\int_{\X}p(x|c)\sum_{a\in\A}\pi_0(a|x)w(c,a)q(a,c,x)dx}\nonumber\\
    &\phantom{\Exp{p(c)}{}}-\Exp{p(c)}{\int_{\X}p(x|c)\sum_{a\in\A}\pi(a|x)q(a,c,x)dx}\nonumber\\
\end{align}
Under Assumption \ref{as4} we have that $\delta^-_{\pi,c,a} \leq \frac{\pi(a | c, x)}{p(a | c, \pi)} \leq \delta^+_{\pi,c,a}, \quad \delta^-_{\pi_0,c,a} \leq \frac{\pi_0(a | c, x)}{p(a | c, \pi_0)} \leq \delta^+_{\pi_0,c,a} \quad \forall x \in \X$ given an action $a \in \A$ and a context $c \in \C$. We denote then $\delta^+_{c,a}=\max\{\delta^+_{\pi,c,a}, \delta^+_{\pi_0,c,a}\}, \delta^-_{c,a}=\min\{\delta^-_{\pi,c,a}, \delta^-_{\pi_0,c,a}\}, \Delta_{a,c}=\delta^+_{c,a} - \delta^-_{c,a}$, and we can give an upper bound as follows:
\begin{align}
    &\Exp{p(c)}{\int_{\X}p(x|c)\sum_{a\in\A}\pi_0(a|x)w(c,a)q(a,c,x)dx}\\ \nonumber
    &\phantom{\Exp{p(c)}{}}-\Exp{p(c)}{\int_{\X}p(x|c)\sum_{a\in\A}\pi(a|x)q(a,c,x)dx}\label{eq:12}\\
    &\leq \Exp{p(c)}{\int_{\X}p(x|c)\sum_{a\in\A}\delta^+_{c,a}p(a|c,\pi_0)w(c,a)q(a,c,x)dx}\\
    &\phantom{\Exp{p(c)}{}}- \Exp{p(c)}{\int_{\X}p(x|c)\sum_{a\in\A}\delta^-_{c,a}p(a|c,\pi)q(a,c,x)dx}\nonumber\\
    &=\Exp{p(c)}{\sum_{a\in\A}p(a|c,\pi)\int_{\X}p(x|c)\delta^+_{c,a}\frac{p(a|c,\pi)}{\cancel{p(a|c,\pi_0)}}\cancel{p(a|c,\pi_0)}q(a,c,x)dx}\nonumber\\
    &\phantom{\Exp{p(c)}{}}-\Exp{p(c)}{\sum_{a\in\A}p(a|c,\pi)\int_{\X}p(x|c)\delta^-_{c,a}q(a,c,x)dx}\nonumber\\
    &=\Exp{p(c)p(a|c,\pi)}{\delta^+_{c,a}\int_{\X}p(x|c)q(a,c,x)dx} - \Exp{p(c)p(a|c,\pi)}{\delta^-_{c,a}\int_{\X}p(x|c)q(a,c,x)dx}\nonumber\\
    &=\Exp{p(c)p(a|c,\pi)}{\Exp{p(x|c)}{q(a,c,x)} (\delta^+_{c,a} - \delta^-_{c,a})}\nonumber\\
    &=\Exp{p(c)p(a|c,\pi)}{\Exp{p(x|c)}{q(a,c,x)} \Delta_{a,c}}\nonumber
\end{align}

Note that in Equation \ref{eq:12} we can follow an analogous path to establish a lower bound:
\begin{align*}
    &\Exp{p(c)}{\int_{\X}p(x|c)\sum_{a\in\A}\pi_0(a|x)w(c,a)q(a,c,x)dx} - \Exp{p(c)}{\int_{\X}p(x|c)\sum_{a\in\A}\pi(a|x)q(a,c,x)dx}\\
    &\geq \Exp{p(c)}{\int_{\X}p(x|c)\sum_{a\in\A}\delta^-_{c,a}p(a|c,\pi_0)w(c,a)q(a,c,x)dx} \nonumber\\
    &\phantom{\Exp{p(c)}{}}-\Exp{p(c)}{\int_{\X}p(x|c)\sum_{a\in\A}\delta^+_{c,a}p(a|c,\pi)q(a,c,x)dx}\nonumber\\
    &=-\Exp{p(c)p(a|c,\pi)}{\Exp{p(x|c)}{q(a,c,x)} \Delta_{a,c}}
\end{align*}

From which we have:
\begin{align*}
 \left\lvert\text{Bias}(\hat{V}_{\text{CHIPS}}(\pi; \D))\right\rvert \leq \left\lvert\Exp{p(c)p(x|c)p(a | c, \pi)}{q(a, c, x)\cdot\Delta_{a,c}}\right\rvert
\end{align*}

\subsection{Proposition \ref{p8}}
\label{apdx:proof-p8}
Since the observations are independent we have that
$$
\begin{aligned}
& N\left(\operatorname{MSE}\left(\hat{V}_{\text {IPS }}(\pi)\right)-\operatorname{MSE}\left(\hat{V}_{\text{CHIPS}}(\pi)\right)\right. \\
& =\mathbb{V}_{x, a, r}[\omega(x, a) r]-\mathbb{V}_{c, a, r}[\omega(a, c) r]-N \operatorname{Bias}\left(\hat{V}_{\text{CHIPS}}(\pi)\right)^2
\end{aligned}
$$
We now analyze the difference in variance:
\begin{align*}
&V_{p(c) p(x \mid c) \pi_0(a \mid x) p(r \mid a, c, x)}[\omega(x, a) r]-V_{p(c) p(x \mid c) \pi_0(a \mid x) p(r \mid a, c, x)}[\omega(a, c) r]\\
&=\mathbb{E}_{p\left(c \mid p(x \mid c) \pi_0(a \mid x) p(r \mid a, c, x)\right.}\left[\omega(x, a) r^2\right]-V(\pi)^2 \\
&\quad-\left(\mathbb{E}_{p(c) p(x \mid c) \pi_b(a \mid x) p(r|a, c, x)}\left[\omega(a, c)^2 \cdot r^2\right]-\left(V(\pi)+\operatorname{Bias}\left(\hat{V}_{\text{CHIPS}}(\pi)\right)\right)^2\right.\\
&=\mathbb{E}_{p(c) p(x \mid c) \pi_0(a \mid x)}\left[\left(\omega(x, a)^2-\omega(a, c)^2\right) \mathbb{E}_{p(r|a, c, x)}\left[r^2\right]\right] \\
&\quad +2 V(\pi) \operatorname{Bias}\left(\hat{V}_{\text{CHIPS}}(\pi)\right)+\operatorname{Bias}\left(\hat{V}_{\text{CHIPS}}(\pi)\right)^2
\end{align*}
This implies that
$$
\begin{aligned}
& N\left(\operatorname{MSE}\left(\hat{V}_{\text {IPS }}(\pi)\right)-\operatorname{MSE}\left(\hat{V}_{\text{CHIPS}}(\pi)\right)\right. \\
& = \mathbb{E}_{p(c) p(x \mid c) \pi_0(a \mid x)}\left[\left(\omega(x, a)^2-\omega(a, c)^2\right) \mathbb{E}_{p(r|a, c, x)}\left[r^2\right]\right] \\
&\quad +2 V(\pi) \operatorname{Bias}\left(\hat{V}_{\text{CHIPS}}(\pi)\right)+(1-N)\operatorname{Bias}\left(\hat{V}_{\text{CHIPS}}(\pi)\right)^2
\end{aligned}
$$

\subsection{Proposition \ref{p4}}
\label{apdx:proof-p4}
\label{proof:p4}
Given the logging policy $\pi_0$ and some evaluation policy $\pi$, the absolute bias of the CHIPS estimator when Assumption \ref{as3}, we have that:
\begin{align}
    \text{Bias}(\hat{V}_{\text{CHIPS}}(V; \D)) &= \Exp{\D}{w(c,a)r} - V(\pi)\nonumber\\
    &=\Exp{p(x)p(c|x)\pi_0(a|x)p(r|a,c,x)}{w(a,c)r} - V(\pi)\nonumber\\
    &=\Exp{p(c)p(x|c)\pi_0(a|x)}{w(a,c)q(a,c)} - \Exp{p(c)p(x|c)\pi(a|x)}{w(a,c)q(a,c)}\nonumber\\
    &=\Exp{p(c)p(x|c)\pi_0(a|x)}{w(a,c)q(a,c)} - \Exp{p(c)p(x|c)\pi(a|x)}{w(a,c)q(a,c)}\nonumber\\
    &=\Exp{p(c)}{\int_{\X}p(x|c)\sum_{a\in\A}\pi_0(a|x)w(c,a)q(a,c)dx} \nonumber\\
    &\phantom{\Exp{p(c)}{}}-\Exp{p(c)}{\int_{\X}p(x|c)\sum_{a\in\A}\pi(a|x)q(a,c)dx}\nonumber\\
    &=\Exp{p(c)}{\sum_{a\in\A}w(c,a)q(a,c)\int_{\X}p(x|c)\pi_0(a|x)dx}\nonumber\\
    &\phantom{\Exp{p(c)}{}}-\Exp{p(c)}{\sum_{a\in\A}q(a,c)\int_{\X}p(x|c)\pi(a|x)dx}\nonumber\\
    &=\Exp{p(c)}{\sum_{a\in\A}w(c,a)q(a,c)p(a|c,\pi_0)} - \Exp{p(c)}{\sum_{a\in\A}q(a,c)p(a|c,\pi)}\nonumber\\
    &=\Exp{p(c)}{\sum_{a\in\mathcal{U}(c,\pi_0)^c}w(c,a)q(a,c)p(a|c,\pi_0)}-\Exp{p(c)}{\sum_{a\in\A}q(a,c)p(a|c,\pi)}\label{eq:13}\\
    &=\Exp{p(c)}{\sum_{a\in\mathcal{U}(c,\pi_0)^c}\frac{p(a|c,\pi_0)}{\cancel{p(a|c,\pi_0)}}\cancel{p(a|c,\pi_0)}q(a,c)} \nonumber\\
    &\phantom{\Exp{p(c)}{}}-\Exp{p(c)}{\sum_{a\in\A}q(a,c)p(a|c,\pi)}\nonumber\\
    &=\Exp{p(c)}{-\sum_{a\in\mathcal{U}(c,\pi_0)}p(a|c,\pi)q(a,c)}\nonumber
\end{align}
Where in Equation \ref{eq:13} we note that $p(a|c,\pi_0) = 0$ if $a \in \mathcal{U}(c,\pi_0)$. Following an analogous procedure we can give an expression for the bias of IPS in a cluster bandits setup:
\begin{align*}
    \text{Bias}(\hat{V}_{\text{IPS}}(V; \D)) &= \Exp{\D}{w(a,x)r} - V(\pi)\nonumber\\
    &=\Exp{p(x)p(c|x)\pi_0(a|x)p(r|a,c,x)}{w(a,x)r} - V(\pi)\nonumber\\
    &=\Exp{p(c)p(x|c)\pi_0(a|x)}{w(a,x)q(a,c)} - \Exp{p(c)p(x|c)\pi(a|x)}{q(a,c)}\nonumber\\
    &=\Exp{p(c)}{\int_{\X}p(x|c)\sum_{a\in\mathcal{U}(c, x, \pi_0)^c}\pi_0(a|x)w(a,x)q(a,c)dx} \nonumber\\
    &\phantom{\Exp{p(c)}{}}-\Exp{p(c)}{\int_{\X}p(x|c)\sum_{a\in\A}\pi(a|x)q(a,c)dx}\nonumber\\
    &=\Exp{p(c)}{\sum_{a\in\mathcal{U}(c, x, \pi_0)^c}q(a,c)\int_{\X}p(x|c)\frac{\pi(a|x)}{\cancel{\pi_0(a|x)}}\cancel{\pi_0(a|x)}dx} \nonumber\\
    &\phantom{\Exp{p(c)}{}}-\Exp{p(c)}{\sum_{a\in\A}q(a,c)\int_{\X}p(x|c)\pi(a|x)dx}\nonumber\\
    &=\Exp{p(c)}{\sum_{a\in\mathcal{U}(c, x, \pi_0)^c}q(a,c)p(a|c,\pi)} - \Exp{p(c)}{\sum_{a\in\A}q(a,c)p(a|c,\pi)}\nonumber\\
    &=\Exp{p(c)}{-\sum_{a\in\mathcal{U}(c, x, \pi_0)}p(a|c,\pi)q(a,c)}\nonumber
\end{align*}

Since $q(a,c) \geq 0$ in the binary reward setting, it follows that $\left\lvert\text{Bias}(\hat{V}_{\text{CHIPS}}(\pi;\D))\right\rvert = \Exp{p(c)}{\sum_{\mathcal{U}(c, \pi_0)} p(a| \pi, c)q(a, c)}$ and $\left\lvert\text{Bias}(\hat{V}_{\text{IPS}}(\pi;\D))\right\rvert = \Exp{p(c)}{\sum_{\mathcal{U}(c, x, \pi_0)} p(a| \pi, c)q(a, c)}$ and consequently we have that:
\begin{align*}
    \left\lvert\text{Bias}(\hat{V}_{\text{IPS}}(\pi;\D))\right\rvert - \left\lvert\text{Bias}(\hat{V}_{\text{CHIPS}}(\pi;\D))\right\rvert &= \Exp{p(c)}{\sum_{\mathcal{U}(c, x, \pi_0)} p(a| \pi, c)q(a, c)} \nonumber\\
    &\phantom{\Exp{p(c)}{}}- \Exp{p(c)}{\sum_{\mathcal{U}(c, \pi_0)} p(a| \pi, c)q(a, c)}\\
    &= \Exp{p(c)}{\sum_{\mathcal{U}(c, x, \pi_0) \setminus \mathcal{U}(c, \pi_0)} p(a| \pi, c)q(a, c)}
\end{align*}

\subsection{Proposition \ref{p9}}
\label{apdx:proof-p9}
Assuming that we have a set of embeddings $e \in \mathcal{E} \subset \mathbb{R}^{d_e}$ associated with the actions $a \in \mathcal{A}$ and an approximation $f_{\phi^*}(r)$ to the importance weights $w(a,x)$:
\begin{equation}
\begin{aligned}
f_{\phi^*}(r):= & \operatorname{argmin} \mathbb{E}_\phi\left[\left(w(a, x)-f_\phi(r)\right)^2\right] \\
& f_\phi \in\left\{f_\phi: \mathbb{R} \rightarrow \mathbb{R} \mid \phi \in \Phi\right\}
\end{aligned}
\end{equation}

Then if we assume that $f_{\phi^*}(r)  = w(a,x) + \epsilon $ for some $\epsilon \in \mathbb{R}$ we have that

\begin{align*}
    &\left|\operatorname{Bias}\left(\hat{V}_{\text{MR}} ; \mathcal{D}\right)\right|-\left|\operatorname{Bias}\left(\hat{V}_{\text {CHIPS }} ; \mathcal{D}\right)\right|\\
    & =-\mathbb{E}_{p(c)}\left[\sum_{a \in \mathcal{U}\left(c, \pi_0\right)} p(a \mid \pi, c) q(a, c)\right]+\operatorname{Bias}\left(\hat{V}_{\text {IPS }} ; D\right)+\mathbb{E}_\mathcal{D}\left[f_{\phi^*}(r) r\right]-\mathbb{E}_{\mathcal{D}}(w(a, x)] \\
    & =-\mathbb{E}_{p(c)}\left[\sum_{a \in \mathcal{U}(c, \pi)} p(a \mid \pi, c) q(a, c)\right]-V(\pi)+\mathbb{E}_\mathcal{D}[f_{\phi^*}(r) r] \\
    & =-\mathbb{E}_{p(c)}\left[\sum_{a \in \mathcal{U}\left( c, \pi_0\right)} p(a \mid \pi, c) q(a, c)\right]-V(\pi)+\mathbb{E}_\mathcal{D}[w(a, x)r]+\epsilon\mathbb{E}_\mathcal{D}[r]\\
    & =-\mathbb{E}_{p(c)}\left[\sum_{a \in \mathcal{U}(c, \pi)} p(a \mid \pi, c) q(a, c)\right] +\mathbb{E}_{p(c)}\left[\sum_{a \in \mathcal{U}(x,c,\pi_0)} q(a, c) \underbrace{\int_{x \in x} p(x \mid c) \pi(a \mid x) dx}_{p(a \mid \pi, c)}\right]\\&\quad\quad-\mathbb{E}_{p(c)}\left[\sum_{a \in \mathcal{A}} q(a, c) \underbrace{\int_{x \in x} p(x \mid c) \pi(a \mid x) d x}_{p(a \mid \pi, c)}\right]+\varepsilon \mathbb{E}_D[r]\\
    &=\mathbb{E}_{p(c)}\left[\sum_{a \in \mathcal{U}(x,c,\pi_0)\setminus\mathcal{U}(c,\pi_0)} q(a, c){p(a \mid \pi, c)}\right]-\mathbb{E}_{p(c)}\left[\sum_{a \in \mathcal{A}} q(a, c)p(a \mid \pi, c)\right]+\varepsilon \mathbb{E}_D[r]\\
    &=-\mathbb{E}_{p(c)}\left[\sum_{a \in (\mathcal{U}(x,c,\pi_0)\setminus\mathcal{U}(c,\pi_0))^c} q(a, c)p(a \mid \pi, c)\right] + \epsilon\mathbb{E}_{p(c)}\left[\sum_{a \in \mathcal{A}} q(a, c)p(a \mid \pi_0, c)\right]
\end{align*}
for the MIPS case, we note that MIPS' bias can also be expressed similarly to CHIPS':

\begin{align*}
& \operatorname{Bias}\left(\hat{V}_{\text {MIPS }} ; D\right)= \\
& =\mathbb{E}_D[w(x, e) r]-V(\pi) \\
& =\mathbb{E}_{p(x) \pi_0(a \mid x) p(e \mid x, a) p(r \mid x, a, e)}[w(x, e) r]-V(\pi) \\
& =\mathbb{E}_{p(x)}\left[\sum_{a \in \mathcal{A}} \pi_0(a \mid x) \sum_{e \in \mathcal{E}} p(e \mid x, a) w(x, e) q(x, e)\right] \\
& \quad\quad-\mathbb{E}_{p(x)}\left[\sum_{a \in \mathcal{A}} \pi(a \mid x) \sum_{e \in \mathcal{E}} p(e \mid x, a) w(x, e) q(x, e)\right] \\
& =\mathbb{E}_{p(x)}\left[\sum_{e \in \mathcal{E}} q(x, e)\left(\sum_{a \in \mathcal{A}} \pi_0(a \mid x), p(e \mid x, a)\right)\right] \\
&\quad\quad -\mathbb{E}_{p(x)}\left[\sum_{e \in \mathcal{E}} q(x, e)\left(\sum_{a \in \mathcal{A}} \pi(a \mid x) p(e \mid x, a)\right)\right] \\
& =E_{p(x)}\left[\sum_{e \in \mathcal{U}\left(e, \pi_0\right)^c} p\left(e \mid x, \pi_0\right) q(x, e) \frac{p(e \mid x, \pi)}{p\left(e \mid x, \pi_0\right)}\right] \\
& \quad\quad-\mathbb{E}_{p(x)}\left[\sum_{e \in \mathcal{E}} q(x, e) p(e \mid x, \pi)\right] \\
& =-\mathbb{E}_{p(x)}\left[\sum_{e \in \mathcal{U}\left(e, \pi_0\right)} p(e \mid x, \pi) q(x, e)\right]
\end{align*}

Therefore the difference in bias is:

\begin{align*}
    &\left|\operatorname{Bias}\left(\hat{V}_{\text{MIPS}} ; \mathcal{D}\right)\right|-\left|\operatorname{Bias}\left(\hat{V}_{\text {CHIPS }} ; \mathcal{D}\right)\right|\\
    & =\mathbb{E}_{p(x)}\left[\sum_{e \in \mathcal{U}\left(e, \pi_0\right)} p(e \mid x, \pi) q(x, e)\right] - \mathbb{E}_{p(c)}\left[\sum_{a \in \mathcal{U}(c, \pi_0)} p(a \mid c, \pi) q(a, c)\right]
\end{align*}

\subsection{Proposition \ref{p6}}
\label{apdx:proof-p6}
\begin{lemma}
    \label{p5}
    Given a policy $\pi$, under Assumption \ref{as2} we have the transformation:
    \begin{equation*}
        w(a,c) = \Exp{\pi_0(x|a, c)}{w(a,x)}
    \end{equation*}
\end{lemma}
\textit{Proof:}

Given a logging policy $\pi_0$ and an evaluation policy $\pi$, in the cluster setting of the bandits problem we have that:
\begin{align}
    w(a,c) &= \frac{p(a|\pi,c)}{p(a|\pi_0,c)} \nonumber\\
        &=\frac{\int_{\X}\pi(a|x)p(x|c)}{p(a|\pi_0,c)}\label{eq:14}\\
        &=\frac{\cancel{p(a|c,\pi_0)}\int_{\X}\frac{\pi(a|x)}{\pi_0(a|x)}\pi_0(x|a,c)}{\cancel{p(a|\pi_0,c)}}\label{eq:15}\\
        &=\Exp{\pi_0(x|a,c)}{w(a,x)}\nonumber
\end{align}
Where we have used the definition $p(a|\pi,c) = \int_{\X}\pi(a|x)p(x|c)$ in Equation \ref{eq:14}, and that $\pi_0(x|a,c)=\frac{p(x|c)\pi_0(a|x)}{p(a|c,\pi_0)}$ in Equation \ref{eq:15}.

Given a logging policy $\pi_0$ and an evaluation policy $\pi$, under Assumption \ref{as2} and Assumption \ref{as3} we have that
\begin{align}
    &N\left(\Var{\D}{\hat{V}_{\text{IPS}}(\pi;\D)} - \Var{\D}{\hat{V}_{\text{CHIPS}}(\pi;\D)}\right)\nonumber\\
    &=N\left(\Var{\D}{\frac{1}{N}\sum_{i=1}^N\frac{\pi(a_i|x_i)}{\pi_0(a_i|x_i)}r_i} - \Var{\D}{\frac{1}{N}\sum_{i=1}^N\frac{p(a_i|c_i,\pi)}{p(a_i|c_i,\pi_0)}r_i}\right)\nonumber\\
    &=\Var{\D}{\frac{\pi(a|x)}{\pi_0(a|x)}r} - \Var{\D}{\frac{p(a|c,\pi)}{p(a|c,\pi_0)}r}\label{eq:16}\\
    &=\left(\Exp{\D}{w(a,x)^2r^2} - \underbrace{\cancelto{0}{\Exp{\D}{w(a,x)r}^2}}_{V(\pi)}\right) - \left(\Exp{\D}{w(a,c)^2r^2} - \underbrace{\cancelto{0}{\Exp{\D}{w(a,c)r}^2}}_{V(\pi)}\right)\label{eq:17}\\
    &=\Exp{p(x)p(c|x)\pi_0(a|x)}{w(a,x)^2\Exp{p(r|a,c,x)}{r^2}} - \Exp{p(x)p(c|x)\pi_0(a|x)}{w(a,c)^2\Exp{p(r|a,c,x)}{r^2}}\nonumber\\
    &=\Exp{p(x)p(c|x)\pi_0(a|x)}{\left(w(a,x)^2 - w(a,c)^2\right)\Exp{p(r|a,c,x)}{r^2}}\nonumber\\
    &=\Exp{p(c)p(x|c)\pi_0(a|x)}{\left(w(a,x)^2 - w(a,c)^2\right)\Exp{p(r|a,c,x)}{r^2}}\nonumber\\
    &=\Exp{p(c)}{\int_{\X}p(x|c)\sum_{a \in \A}\pi_0(a|x)\left(w(a,x)^2 - w(a,c)^2\right)\Exp{p(r|a,c,x)}{r^2}dx}\nonumber\\
    &=\Exp{p(c)}{\sum_{a\in\A}\int_{\X}p(x|c)\pi_0(a|x)\left(w(a,x)^2 - w(a,c)^2\right)\Exp{p(r|a,c,x)}{r^2}dx}\nonumber\\
    &=\Exp{p(c)}{\sum_{a\in\A}\int_{\X}\frac{\pi_0(x|a,c)p(a|c,\pi_0)}{\cancel{\pi_0(a|x)}}\cancel{\pi_0(a|x)}\left(w(a,x)^2 - w(a,c)^2\right)\Exp{p(r|a,c,x)}{r^2}dx}\label{eq:18}\\
    &=\Exp{p(c)}{\sum_{a\in\A}p(a|c,\pi_0)\int_{\X}\pi_0(x|a,c)\left(w(a,x)^2 - w(a,c)^2\right)\Exp{p(r|a,c,x}{r^2}dx}\nonumber\\
    &=\Exp{p(c)p(a|c,\pi_0)}{\left(\int_{X}\pi_0(x|a,c)w(a,x)^2dx - w(a,c)^2\cancelto{1}{\int_{\X}\pi_0(x|a,c)dx}\right)\Exp{p(r|a,c,x}{r^2}}\nonumber\\
    &=\Exp{p(c)p(a|c,\pi_0)}{\left(\Exp{\pi_0(x|a,c)}{w(a,x)^2} - \Exp{\pi_0(x|a,c)}{w(a,x)}^2\right)\Exp{p(r|a,c,x}{r^2}}\label{eq:19}\\
    &=\Exp{p(c)p(a|c,\pi_0)}{\Var{\pi_0(x|a,c)}{w(a,x)}\Exp{p(r|a,c,x)}{r^2}} \geq 0\nonumber
\end{align}
Note in Equation \ref{eq:16} we used that the samples in $\D$ are i.i.d, in particular the linearity of variance under this condition. The cancellation of terms in Equation \ref{eq:17} results from IPS and CHIPS being unbiased under Assumptions \ref{as2} and \ref{as3}. In Equation \ref{eq:18} we used that $\pi_0(x|a,c)=\frac{p(x|c)\pi_0(a|x)}{p(a|c,\pi_0)}$, while Equation \ref{eq:19} uses \cref{p5}.

\subsection{Proposition \ref{p7}}
\label{apdx:proof-p7}

The first thing we need to note is that CHIPS and MIPS are in different spaces regarding the contextual bandits generating process. MIPS assumes the existence of an action embedding space $e \in \mathcal{E} \subseteq \mathbb{R}^{d_e}$ and CHIPS assumes the existence of a partition of the context space $\mathcal{C}:=\left\{\mathcal{C}_i\right\}_{i=1}^K$ with $\mathcal{C}_i \subset \mathcal{X}$ and $c_i \cap c_j=\varnothing$. For joining this spaces, we assume that given a policy $\pi$, at every iteration of the data generation process, apart from the classical context ($x\in\mathcal{X}$), action ($a \in \mathcal{A}$) and reward ($r \in [0, r_{max}]\subset\mathbb{R})$, we observe a cluster $c \sim p(c \mid x)$ and an action embedding $e \sim p(e \mid a, c, x)$. Given a policy $\pi$ the policy value $V(\pi)$ equation can be then refined to: 
\begin{align*}
    V(\pi)&:= \mathbb{E}_{p(x) p(c \mid x) \pi(a \mid x) p(e \mid a, c, x) p(r \mid e, a, c, x)[r]}\\
    &=\mathbb{E}_{p(x) p(c \mid x) \pi(a \mid x) p(e \mid a, c, x) q(e,a,c,x)}
\end{align*}
Here $q(e, a, c, x):=\mathbb{E}_{p(r \mid e, a, c, x)}[r]$. Note that as in MIPS and CHIPS case, the refinement does not contradict the classical policy value definition.

We also need to refine $p(a \mid c, \pi)$ (from CHIPS) and $p(e \mid a, \pi)$ (from MIPS) in the joint space:

\begin{align*}
    &p(a \mid c, \pi)=\sum_{e \in \varepsilon} \int_\chi p(e \mid a, c, x)p(x \mid c)\pi(a \mid x)\\
    &p(e \mid x, \pi):=\sum_{c \in \mathcal{C}}\sum_{a \in \mathcal{A}} p(e \mid a,c,x) p(c \mid x)\pi(a \mid x) 
\end{align*}
It is important to note that after joining the context space, to make a fair comparison between MIPS and CHIPS, there are some dependencies that we want to eliminate to prevent information from passing between variables that were not originally in the definition of MIPS and CHIPS. In particular, for CHIPS, we eliminate the dependency of the cluster with respect to the embedding given the context and the action (i.e., $c \bot e \mid (x,a)$), and for MIPS, the dependency of the action with respect to the cluster given the embedding and the context (i.e., $a \bot c \mid (x,e)$). From \cref{p6} we know that $\Var{\D}{\hat{V}_{\text{IPS}}(\pi;\D)} \geq \Var{\D}{\hat{V}_{\text{CHIPS}}(\pi;\D)}$ and from MIPS Theorem 3.6 we know that $\Var{\D}{\hat{V}_{\text{IPS}}(\pi;\D)} \geq \Var{\D}{\hat{V}_{\text{MIPS}}(\pi;\D)}$. Therefore, we need to make a comparison between $\Var{\D}{\hat{V}_{\text{MIPS}}(\pi;\D)}$ and $\Var{\D}{\hat{V}_{\text{CHIPS}}(\pi;\D)}$.

To follow the structure of \cref{p6}, we are going to assume that Assumptions 3.1 and 3.2 hold as well as their counterparts from MIPS. The following identities hold under these conditions:

\begin{align*}
\centering
    &p(x \mid c)\pi(a \mid x) = \frac{p(e \mid x, \pi)p(c\mid x,a)\pi_0(a \mid x)p(x)}{p(e \mid x, \pi_0)p(c)}\\
    &p(e \mid x, a, c)=\frac{p(x \mid e, a, c) p(e \mid a, c) p\left(a \mid c, \pi_0\right)}{p(c \mid x, a) \pi_0(a \mid x) p(x)}\\
\end{align*}

Now, under these conditions, we need a relation between the weights of MIPS and CHIPS:
\begin{align*}
    \centering
    \omega(a, c)^2&=\frac{p(a \mid c, \pi)}{p\left(a \mid c, \pi_0\right)}\\
    &=\frac{\int_{\mathcal{X}} p(x \mid c) \sum_{e \in \mathcal{E}} \pi(a \mid x) p(e \mid c, a, x)}{p\left(a \mid c, \pi_0\right)}\\
    &=\frac{\int_x \sum_{e \in \mathcal{E}} w(e, x) p(x \mid e, a, c) p(e|a, c) p(a \mid c, \pi_0)}{p(a \mid c, \pi_0)}\\
    &=\sum_{e \in \varepsilon} p(e|a, c) \int_x p(x \mid e, a, c) \omega(e, x)\\
    &=\mathbb{E}_{p(e \mid a, c) p(x \mid e, a, c)}[\omega(e, x)]
\end{align*}

Therefore the scaled difference in variance can be expressed as:
\begin{align*}
&N\left(\mathbb{V}_{\mathcal{D}}\left[\hat{V}_{\mathrm{MIPS}}(\pi ; \mathcal{D})\right]-\mathbb{V}_{\mathcal{D}}\left[\hat{V}_{\mathrm{CHIPS}}(\pi ; \mathcal{D})\right]\right)\\
& =N\left(\mathbb{V}_{\mathcal{D}}\left[\frac{1}{N} \sum_{i=1}^N \frac{\pi\left(e_i \mid x_i\right)}{\pi_0\left(e_i \mid x_i\right)} r_i\right]-\mathbb{V}_{\mathcal{D}}\left[\frac{1}{N} \sum_{i=1}^N \frac{p\left(a_i \mid c_i, \pi\right)}{p\left(a_i \mid c_i, \pi_0\right)} r_i\right]\right) \\
& =\mathbb{V}_{\mathcal{D}}\left[\frac{\pi(e \mid x)}{\pi_0(e \mid x)} r\right]-\mathbb{V}_{\mathcal{D}}\left[\frac{p(a \mid c, \pi)}{p\left(a \mid c, \pi_0\right)} r\right] \\
& =(\mathbb{E}_{\mathcal{D}}\left[\omega(e, x)^2 r^2\right]-\underbrace{\mathbb{E}_{\mathcal{D}}[\omega(e, x) r]^2}_{V(\pi)} )-(\mathbb{E}_{\mathcal{D}}\left[\omega(a, c)^2 r^2\right]-\underbrace{\mathbb{E}_{\mathcal{D}}[\omega(a, c) r]^2}_{V(\pi)})\\
&=\mathbb{E}_{p(c)}\left[\int_{\mathcal{X}} p(x \mid c) \sum_{a \in \mathcal{A}} \pi_0(a \mid x) \sum_{e \in \mathcal{E}} p(e \mid a, c, x)\left(\omega(e, x)^2-\omega(a, c)^2\right) \mathbb{E}_{p(r \mid a, c)}\left[r^2\right] dx\right]\\
&=\mathbb{E}_{p(c)}\left[\sum_{a \in A} p\left(a \mid c, \pi_0\right) \sum_{e \in \mathcal{E}} p(e \mid a, c) \int_{\mathcal{X}} p(x \mid e, a, c)\left(\omega(e, x)^2-\omega^2(a, c)\right) \mathbb{E}_{p(r|a,c)}\left[r^2\right] dx\right]\\
&=\mathbb{E}_{p(c) p\left(a \mid c, \pi_0\right)}\left[\mathbb{E}_{p(r \mid a, c)}\left[r^2\right]\left(\mathbb{E}_{p(e \mid a, c) p(x \mid e, a, c)^2}\left[\omega(e, x)^2\right]\right)\right]\\
&\phantom{{\mathbb{E}_{p(c) p\left(a \mid c, \pi_0\right)}}{}} - \mathbb{E}_{p(c) p\left(a \mid c, \pi_0\right)}\left[\mathbb{E}_{p(r \mid a, c)}\left[r^2\right]\left(\omega(a, c)^2 \sum_{e \in \mathcal{E}} p(e \mid a, c) \int_{\mathcal{X}} p(x \mid e, a, c) d x\right)\right]\\
&=\mathbb{E}_{p(c) p\left(a \mid c, \pi_0\right)}\left[\mathbb{E}_{p(r \mid a, c)}\left[r^2\right]\left(\mathbb{E}_{p(e \mid a, c) p(x \mid e, a, c)}\left[\omega(e, x)^2\right]-\mathbb{E}_{p(e \mid a, c) p(x \mid e, a, c)}\left[\omega(e, x)]^2\right.\right)\right]\\
&=\mathbb{E}_{p(c) p\left(a \mid c, \pi_0\right)}\left[\mathbb{E}_{p\left(r \mid a, c\right)}\left[r^2\right] \mathbb{V}_{p(e \mid a, c) p(x \mid e, a, c)}[\omega(e, x)]\right] \geq 0
\end{align*}

This implies that under Assumptions \ref{as2} and \ref{as3} (and their counterparts in MIPS), the variance of CHIPS is lower than the variance of MIPS, proving the proposition.

\section{REWARD ESTIMATES DERIVATION}
\label{apdx:reward-estimates}
\subsection{MAP}
From the setting in Subsection \ref{subs:3.2} we denote $R_c:=\{r_i\}_{i=1}^M$ as the rewards observed in cluster $c$ from the logging data\footnote{Here we refer to the already transformed version using clusters. See definition of $\tau$ in Subsection \ref{subs:3.2}}. We consider $R_c$ as independent trials of a Bernoulli random variable with parameter $\theta$ (i.e., $R_c \stackrel{i.i.d.}{\sim} \text{Ber}(\theta)$). Therefore, we have that the likelihood can be expressed as:
\begin{align*}
    p(R_c|\theta) &= \prod_{i=1}^M p(r_i|\theta)\\
    &= \prod_{i=1}^M \theta^{r_i}(1-\theta)^{1-r_i}\\
    &= \theta^{\sum_{i=1}^Mr_i}(1-\theta)^{M-\sum_{i=1}^M}
\end{align*}
Using a Beta distribution as a prior we have that:
\begin{align*}
    p(\theta) = \text{Beta}(\theta|\alpha,\hat{\beta}) = \frac{1}{\mathcal{B}(\alpha,\hat{\beta})}\theta^{\alpha-1}(1-\theta)^{\hat{\beta}-1}
\end{align*}

Where $\mathcal{B}(\alpha,\hat{\beta}) = \frac{\Gamma(\alpha)\Gamma(\hat{\beta})}{\Gamma(\alpha+\hat{\beta})}$ and $\Gamma(\cdot)$ is the Gamma function. The posterior probability can then be expressed as:
\begin{align*}
    p(\theta|R_c) &\propto p(R_c|\theta)p(\theta)\\
    &\propto\theta^{\sum_{i=1}^Mr_i}(1-\theta)^{M-\sum_{i=1}^M}\frac{1}{\mathcal{B}(\alpha,\hat{\beta})}\theta^{\alpha-1}(1-\theta)^{\hat{\beta}-1}\\
    &\propto\theta^{\alpha-1+\sum_{i=1}^Mr_i}(1-\theta)^{\hat{\beta}-1+M-\sum_{i=1}^Mr_i}\\
    &\propto \text{Beta}\left(\theta \mkern5mu\bigg|\mkern5mu \alpha+\sum_{i=1}^Mr_i,\mkern5mu \hat{\beta}+M-\sum_{i=1}^Mr_i\right)
\end{align*}
The MAP estimator of $\theta$ is the mode of the resulting Beta distribution, i.e.
\begin{equation*}
    \hat{\theta}_{\text{MAP}} = \frac{(\alpha - 1) + \sum_{i=1}^Mr_i}{\alpha + \hat{\beta} + M - 2}
\end{equation*}
\subsection{ML}
Using the same setting as in the previous section ($R_c \stackrel{i.i.d.}{\sim} \text{Ber}(\theta)$) we have that the maximum likelihood estimation can be expressed as 
\begin{align*}
    \hat{\theta}_{\text{ML}} &= \mathop{\arg \max}\limits_{\theta \in \Theta}\left\{{\prod_{i=1}^M \theta^{r_i}(1-\theta)^{1-r_i}}\right\}\\
    &= \mathop{\arg \max}\limits_{\theta \in \Theta}\left\{\underbrace{{\log(\theta)\cdot\sum_{i=1}^M r_i + \log((1-\theta))\cdot\sum_{i=1}^M(1-r_i)}}_{l(\theta)}\right\}
\end{align*}
We now search for local maxima by setting the differential to 0:
\begin{align*}
    \frac{\partial l(\theta)}{\partial \theta} = 0 &\implies\frac{\sum_{i=1}^M r_i}{\theta} + \frac{\sum_{i=1}^M (1-r_i)}{(1-\theta)} = 0\\
    &\implies \sum_{i=1}^M r_i - \theta\sum_{i=1}^M r_i = \theta\sum_{i=1}^M (1-r_i)\\
    &\implies \hat{\theta}_{\text{ML}} = \frac{1}{M}\sum_{i=1}^M r_i
\end{align*}

\section{EXPERIMENTAL PARAMETERS AND HARDWARE}
\label{apdx:params}
\begin{table}[ht]
\begin{tabular}{|c|c|l|}
\hline
\textbf{Parameter} & \textbf{Value} & \multicolumn{1}{c|}{\textbf{Description}}                                                     \\ \hline
$c_{exp}$             & 10             & Radius of the n-dimensional ball for context space generation.                  \\ \hline
$c_{rad}$             & 1              & Cluster generation radius.                                                                    \\ \hline
$d_x$             & 2              & Dimension of context vectors.                                                                 \\ \hline
$x_{num}$             & 1.000          & No. of different context vectors in the experiment.                                        \\ \hline
$a_{num}$             & 10             & No. of actions in the experiment.                                                          \\ \hline
$c_{num}$             & 10             & No. of clusters in the experiment.                                                         \\ \hline
$n_{samples}$         & 50.000         & No. of logged samples to use in the experiment.                                            \\ \hline
$emp_{c\_num}$        & 100            & No. of clusters to use empirically by the clustering method.              \\ \hline
$e_{len}$             & 1.000.000      & No. of samples extracted from the dataset for the evaluation policy                        \\ \hline
$b_{len}$             & 1.000.000      & No. of samples extracted from the dataset for the evaluation policy                        \\ \hline
$\sigma$              & 0.2            & Context-specific behaviour deviation from cluster behaviour.                                  \\ \hline
$\beta$               & -1             & Deviation between evaluation and logging policies. \\ \hline
$\alpha$                   & 20             & Parameter from beta distribution in Bayesian inference \\ \hline
$\hat{\beta}$                   & 20             & Parameter from beta distribution in Bayesian inference \\ \hline
\end{tabular}
\caption{Parameters used in the basic configuration for experiments for generation and estimation.}
\end{table}

\begin{table}[h!]
\begin{tabular}{|l|l|}
\hline
CPU   & AMD Ryzen Threadripper PRO 3975WX \\ \hline
RAM   & 256 GB                            \\ \hline
Cores & 64                                \\ \hline
GPU   & 2x Nvidia A100 160GB                   \\ \hline
\end{tabular}
\label{tab:resources}
\caption{Specifications of the machine in which the experiments were executed.}
\end{table}

\section{ADDITIONAL EXPERIMENTS}
\label{apdx:extra}
\subsection{Synthetic Experiments}
\label{apdx:synth-extra}
\begin{figure}[ht]
    \centering
    \includegraphics[width=\textwidth]{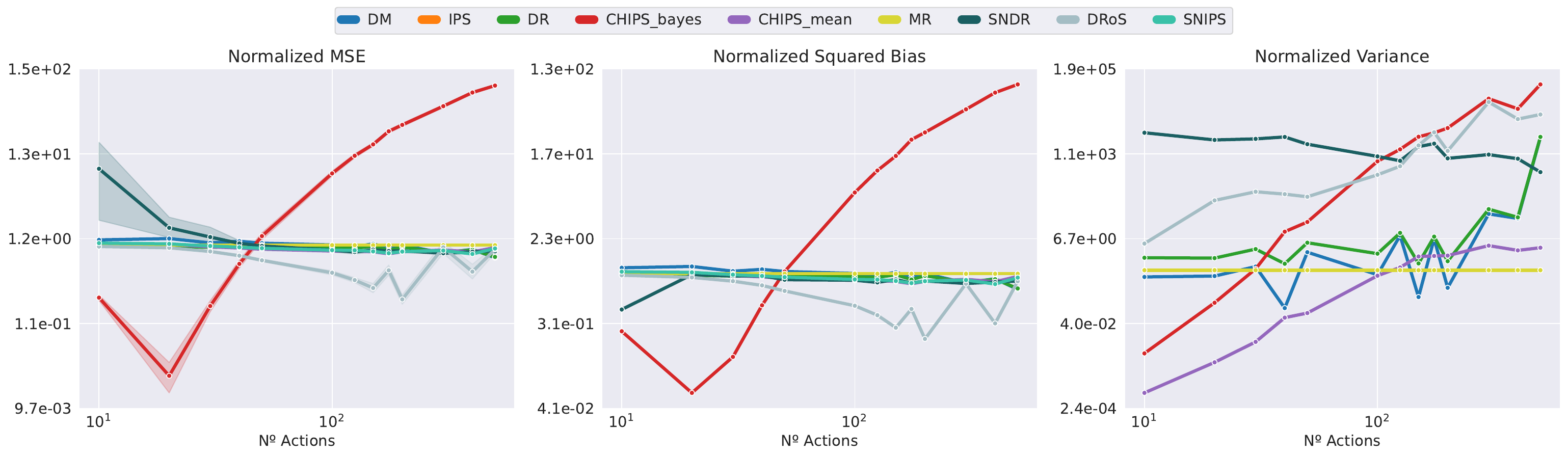}
    \caption{From left to right, MSE, Bias, and Variance of the CHIPS estimator compared to baselines while varying the number of actions.}
    \label{fig:actions}
\end{figure}

\textbf{Number of actions.} From the fixed basic configuration that uses 100 clusters for CHIPS' estimates, we observe a progressive deterioration in the estimator capabilities when increasing the number of actions (see \cref{fig:actions}). We theorize that this behaviour might be a consequence of the violation of Assumption \ref{as2} when trying to group contexts using an excessive number of clusters in a large action space, resulting in deficient actions inside the clusters. This problem can be mitigated by decreasing the number of clusters used in the clustering method for the CHIPS estimation (see \cref{fig:bivariate} (left)).

\textbf{Number of samples.} We observe an approximation to the performance of IPS as we increase the number of samples in the logged data that we identify as an effect of reducing the number of observed deficient action-context pairs in IPS, converging to an unbiased estimator under Assumption \ref{as1} (see \cref{fig:syn} (right)). In this case, the clustering effects under CHIPS become less noticeable according to \cref{cor:1} since $\mathcal{U}(c, x, \pi_0) \setminus \mathcal{U}(c, \pi_0) \rightarrow \emptyset$. It is worth mentioning that increasing the number of clusters when enough samples are available, as well as reducing it in the opposite case, can improve the performance of the CHIPS estimates, as shown in \cref{fig:bivariate} (right).

\begin{figure}[ht]
    \centering
    \includegraphics[width=\textwidth]{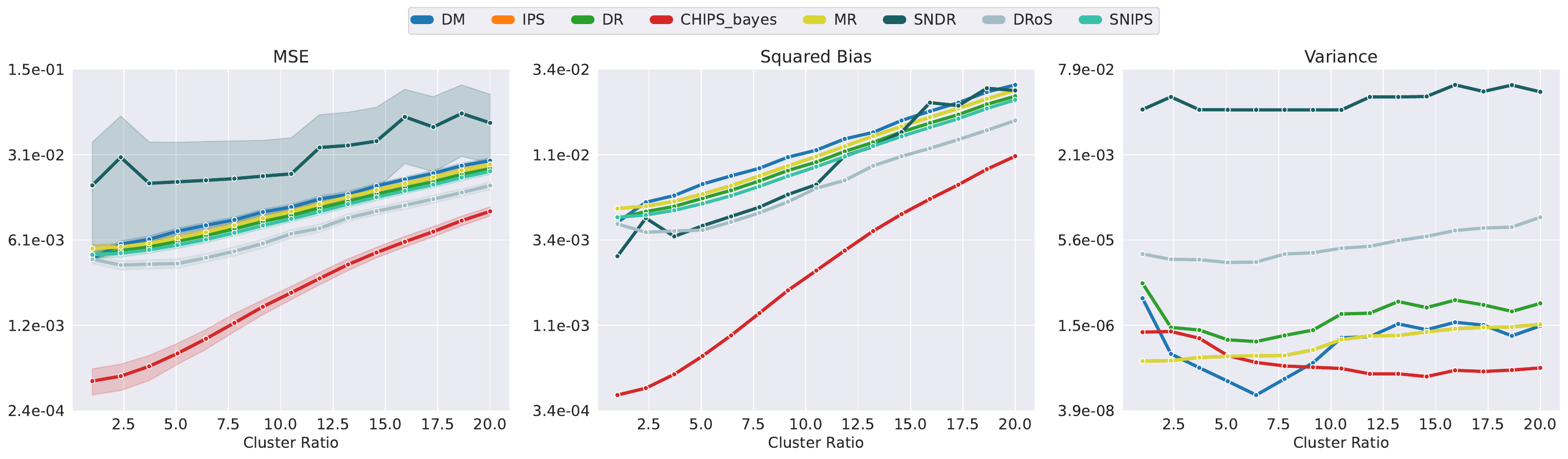}
    \caption{From left to right, MSE, Bias, and Variance of the CHIPS estimator compared to baselines while varying the radius of the clusters generated.}
    \label{fig:rad}
\end{figure}

\textbf{Cluster radius.} Increasing the cluster radius in the generation process affects the separability of the cluster space and complicates the partitioning in clusters complying with Assumption \ref{as3}. In this case, we could find significant differences in context behaviour for both actions and rewards within a cluster, resulting in increased bias from the empirical approximations. Therefore, we observe a convergence to IPS' performance as cluster radius increases since the context space becomes less separable (see \cref{fig:rad}).

\begin{figure}[ht]
    \centering
    \includegraphics[width=\textwidth]{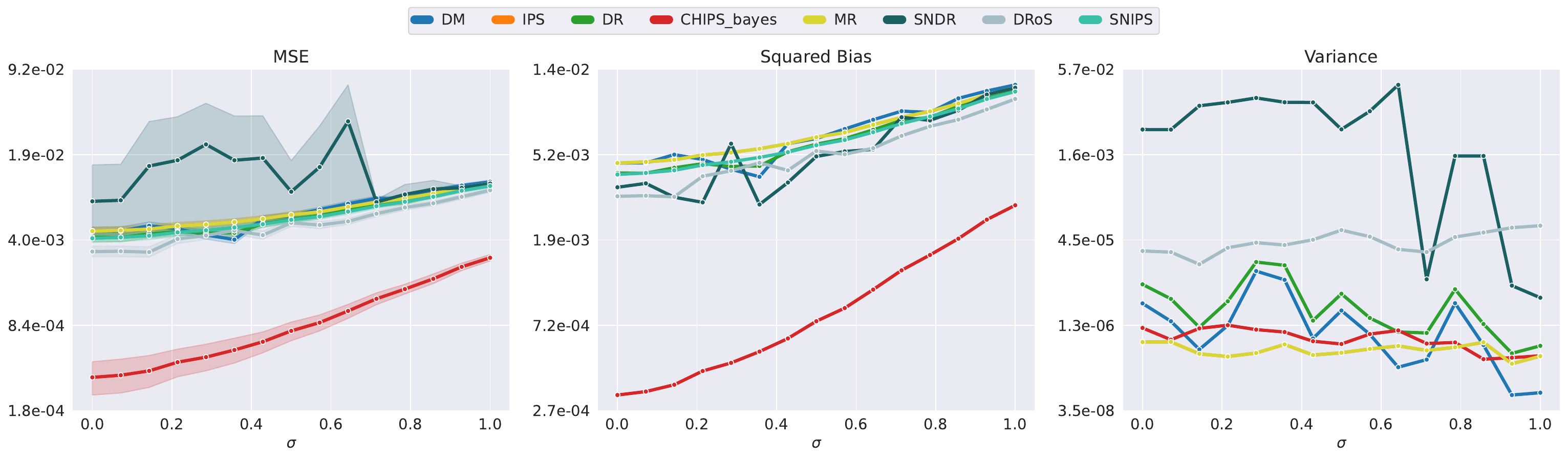}
    \caption{From left to right, MSE, Bias, and Variance of the CHIPS estimator compared to baselines while varying the context-specific noise $\sigma$.}
    \label{fig:sigma}
\end{figure}

\textbf{Sigma.} Increasing context-specific noise in the generation process produces a similar effect as in the cluster radius case. In particular, the larger the noise, the more common it is to observe inconsistent behaviour in actions and rewards for contexts within a cluster, complicating the approximation of a homogeneous cluster-wise behaviour and resulting again in a bias increase (see \cref{fig:sigma}). 

\textbf{Alpha (prior).} In this experiment, we vary the alpha parameter of the Beta prior maintaining all other settings fixed. Like in the number of clusters case, we observe a similar v-shaped graph indicating that, as expected from the previous $\beta$ analysis (see \cref{sec:synthetic results}), the CHIPS (MAP) estimator is sensitive to the prior. In particular, lower values push the expected reward of each cluster to the ML’s estimate, while higher values push it to the prior’s expected value, decreasing performance in both cases (see \cref{fig:syn} center). For different values of distributional shift ($\beta$), the optimal value will depend on the \emph{resistance} MAP offers to converge to the ML estimate, favouring lower values as $\beta$ becomes larger (see \cref{fig:alpha_beta}). 

\begin{figure}[ht]
    \centering
    \includegraphics[width=0.8\textwidth]{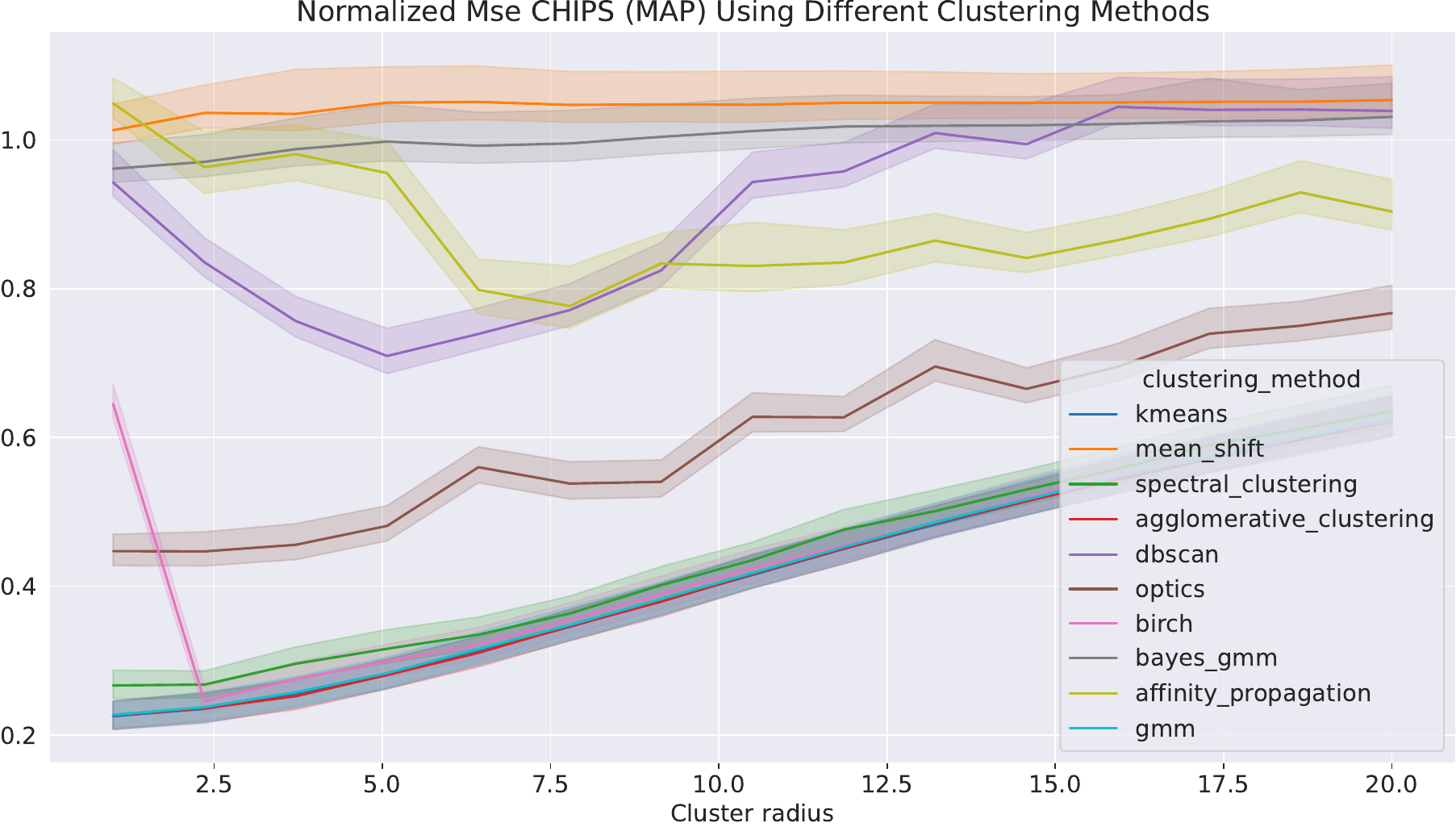}
    \caption{Normalized MSE of CHIPS (MAP) using different clustering methods with respect to IPS.}
    \label{fig:clustering-methods}
\end{figure}

\textbf{Clustering Method.} In this experiment, we evaluate the performance of the CHIPS (MAP) estimator using different clustering methods while varying the clustering radius in the synthetic generation process. In \cref{fig:clustering-methods}, we observe that using Mean Shift \citep{mean-shift} or Bayesian Gaussian Mixture \citep{bishop, variational-bayesian, variational-inference} fails to separate the context space resulting in the same performance as IPS. DBSCAN \citep{dbscan} mitigates IPS' increase in mean squared error when the context space is easier to separate (i.e., lower radii values) but converges to IPS when the context is complicated to separate (i.e., higher radii values). Affinity Propagation \citep{affinity-propagation} follows a similar behaviour to DBSCAN but still offers some improvement with respect to IPS when the space is difficult to separate. OPTICS \citep{optics} makes a general improvement to the Affinity Propagation performance, especially noticeable when the context space is separable. MiniBatch K-Means \citep{kmeans}, Gaussian Mixture \citep{gmm}, Birch \citep{birch}, Spectral Clustering \citep{spectral}, and Agglomerative Clustering \citep{hierarchical} have similar performance, outperforming Affinity Propagation for the separable case. We also note a general upward tendency in mean squared error for every clustering method as the space becomes more complicated to separate.
 
\begin{figure}[ht]
    \centering
    \includegraphics[width=\textwidth]{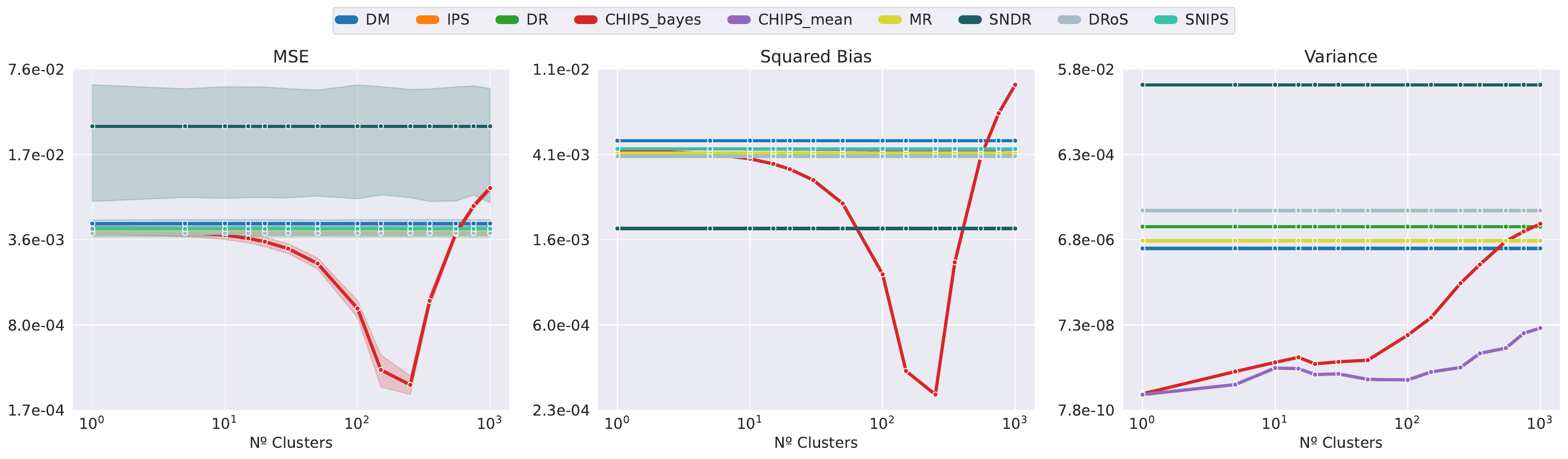}
    \caption{From left to right, MSE, Bias, and Variance of the CHIPS estimator compared to baselines while varying the number of clusters.}
    \label{fig:clusters}
\end{figure}

\begin{figure}[ht]
    \centering
    \includegraphics[width=\textwidth]{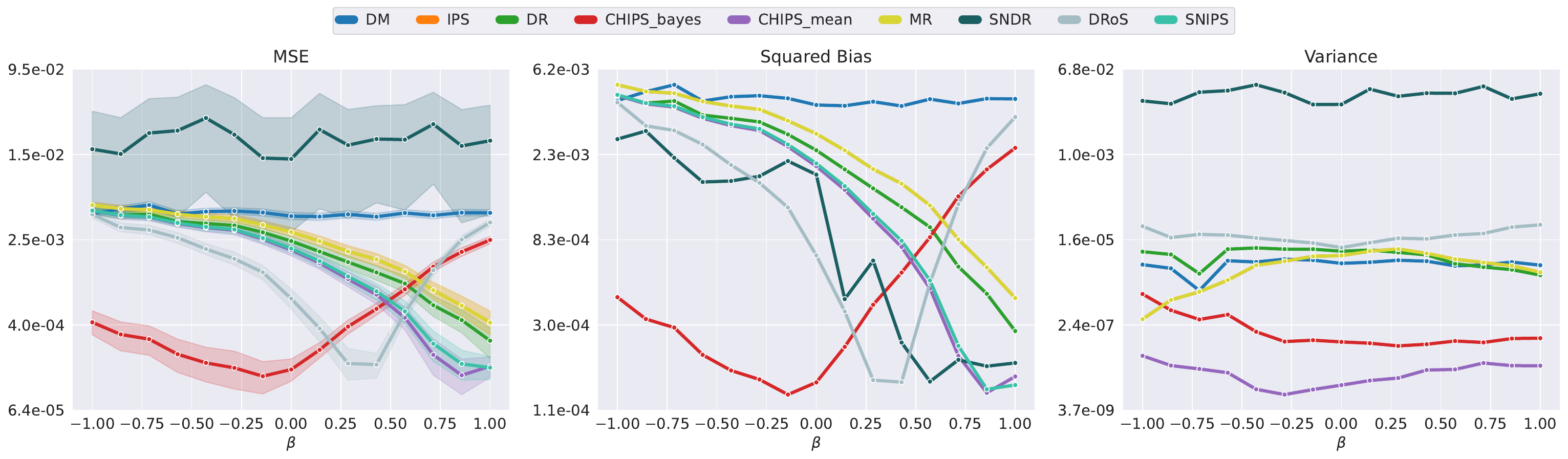}
    \caption[ftn]{From left to right, MSE, Bias, and Variance of the CHIPS estimator compared to baselines while varying $\beta$ values.\footnotemark}
    \label{fig:beta}
\end{figure}
\footnotetext{In this case we used a slightly different version of the configuration settings to make a more challenging environment in which we use 10.000 samples and consequently reduce the number of empirical cluster estimation to 30 to easily assess the role that similarity of logging and evaluation policies play in CHIPS capabilities.}

\begin{figure}[ht]
    \centering
    \includegraphics[width=\textwidth]{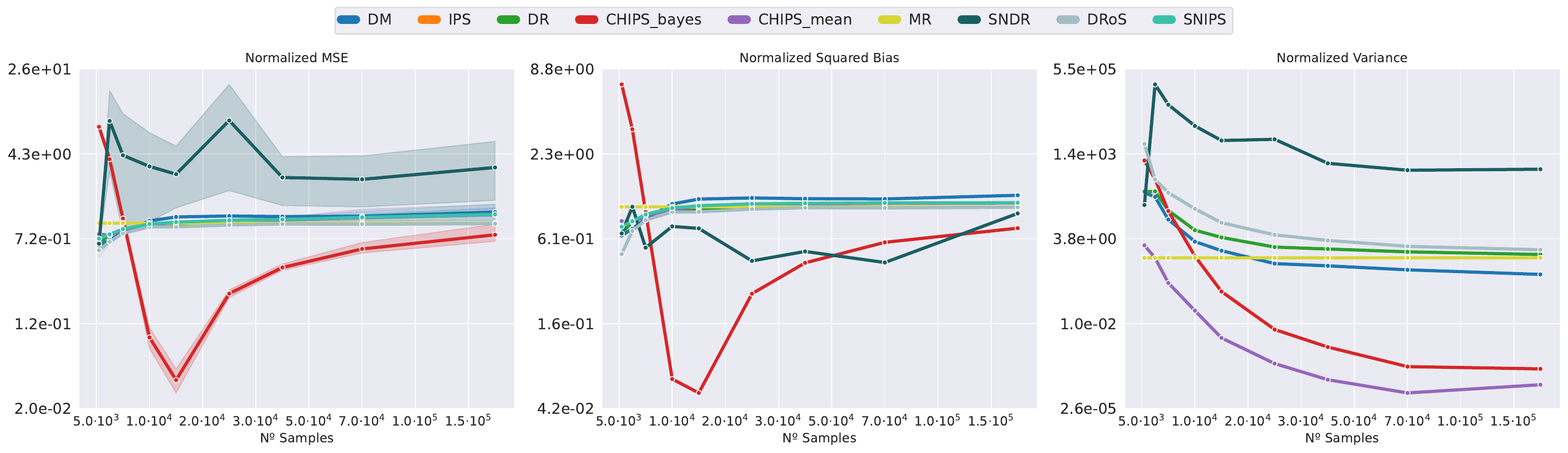}
    \caption{From left to right, MSE, Bias, and Variance of the CHIPS estimator compared to baselines while varying the number of samples provided from the logging policy.}
    \label{fig:samples}
\end{figure}

\begin{figure}[ht]
    \centering
    \includegraphics[width=\textwidth]{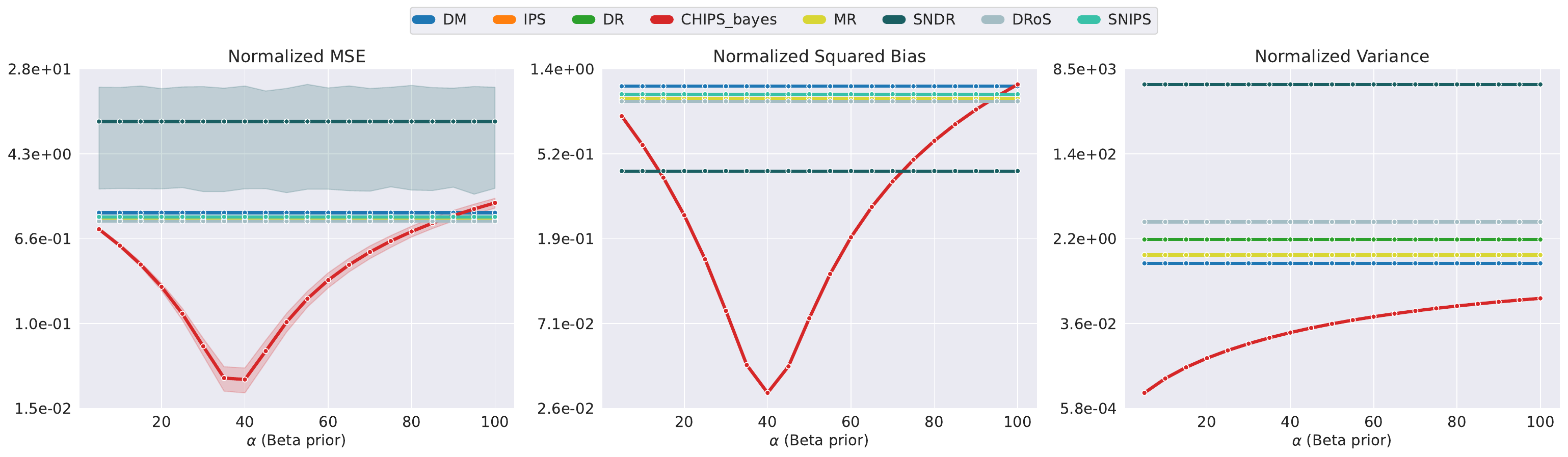}
    \caption{From left to right, MSE, Bias, and Variance of the CHIPS estimator compared to baselines while varying the $\alpha$ parameter.}
    \label{fig:alpha_bayes}
\end{figure}

\begin{figure}[ht]
    \centering
    \includegraphics[width=\textwidth]{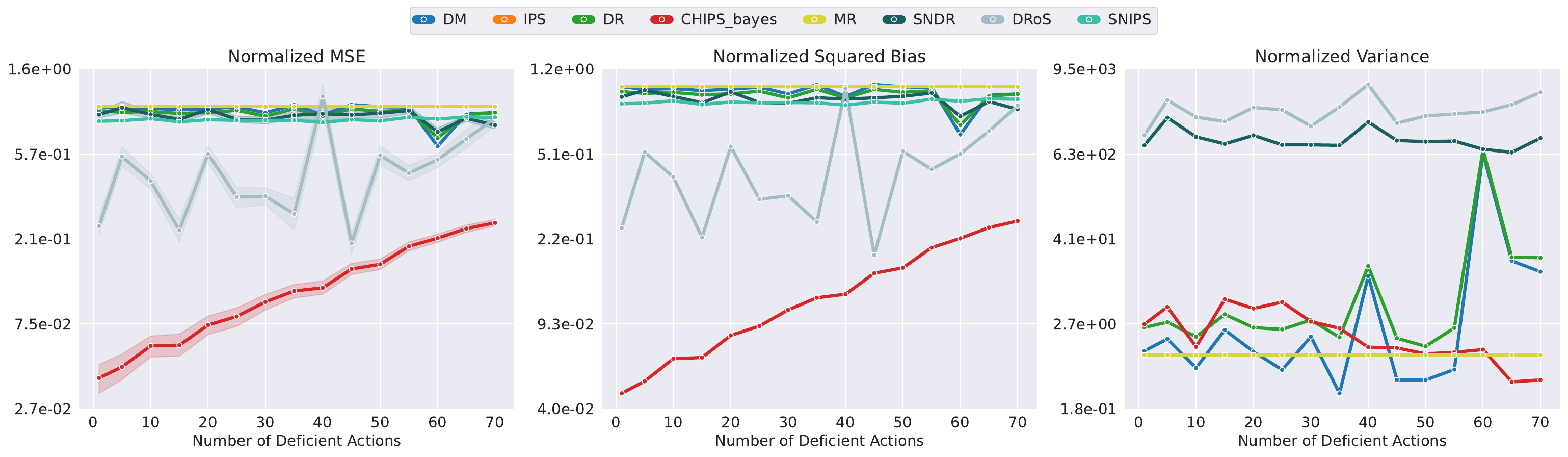}
    \caption{From left to right, MSE, Bias, and Variance of the CHIPS estimator compared to baselines while varying the number of deficient actions.}
    \label{fig:n_deficient}
\end{figure}
\subsubsection{Bi-parametric variations}
\label{apdx:bi}
\begin{figure}[ht]%
    \centering
    \subfloat[\centering Normalized performance of CHIPS (MAP) with respect to IPS using different number of clusters and actions.]{{\includegraphics[width=0.85\textwidth]{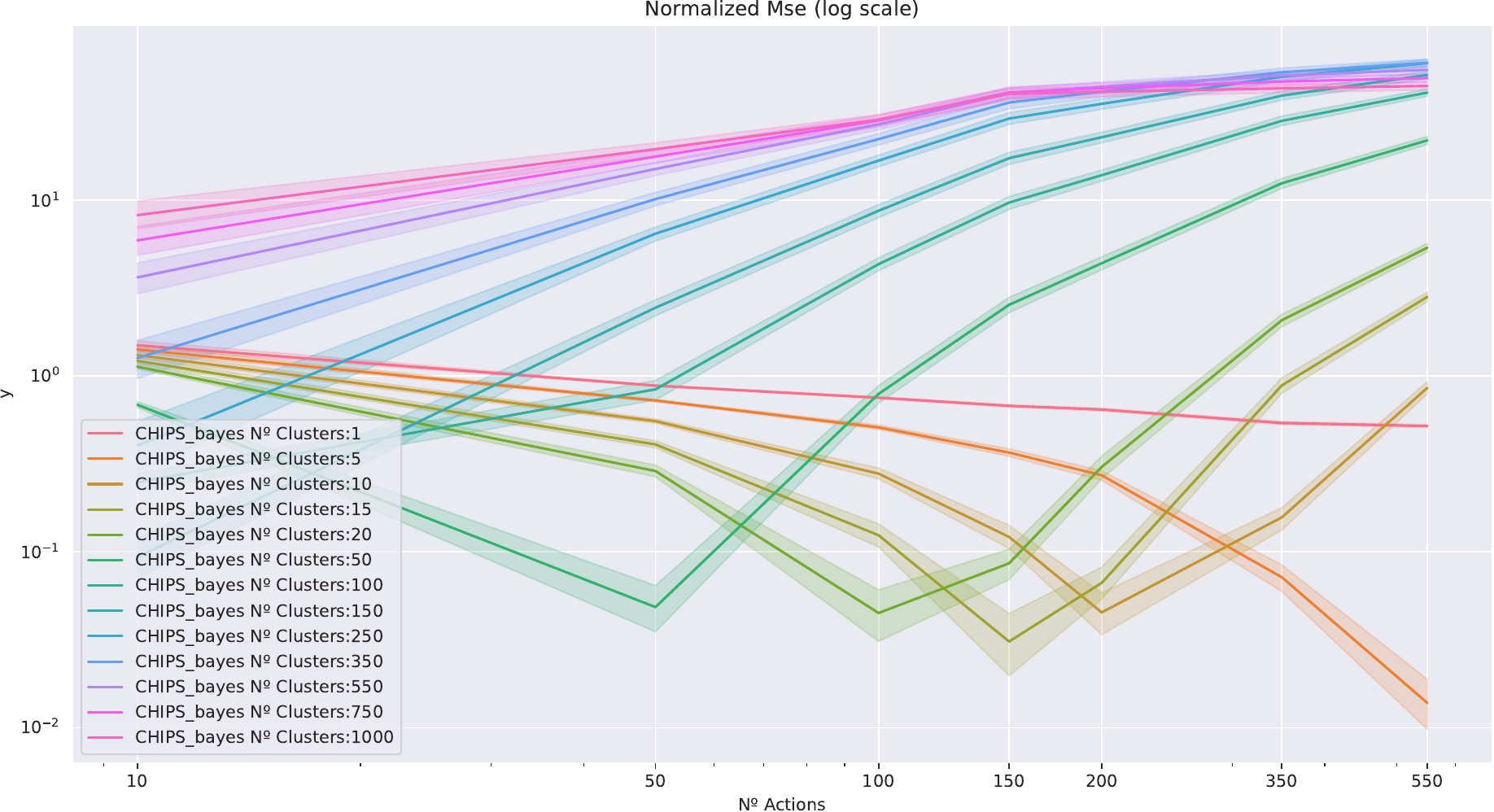} }}%
    \qquad
    \subfloat[\centering Normalized performance of CHIPS (MAP) with respect to IPS using different number of clusters and logging samples.]{{\includegraphics[width=0.85\textwidth]{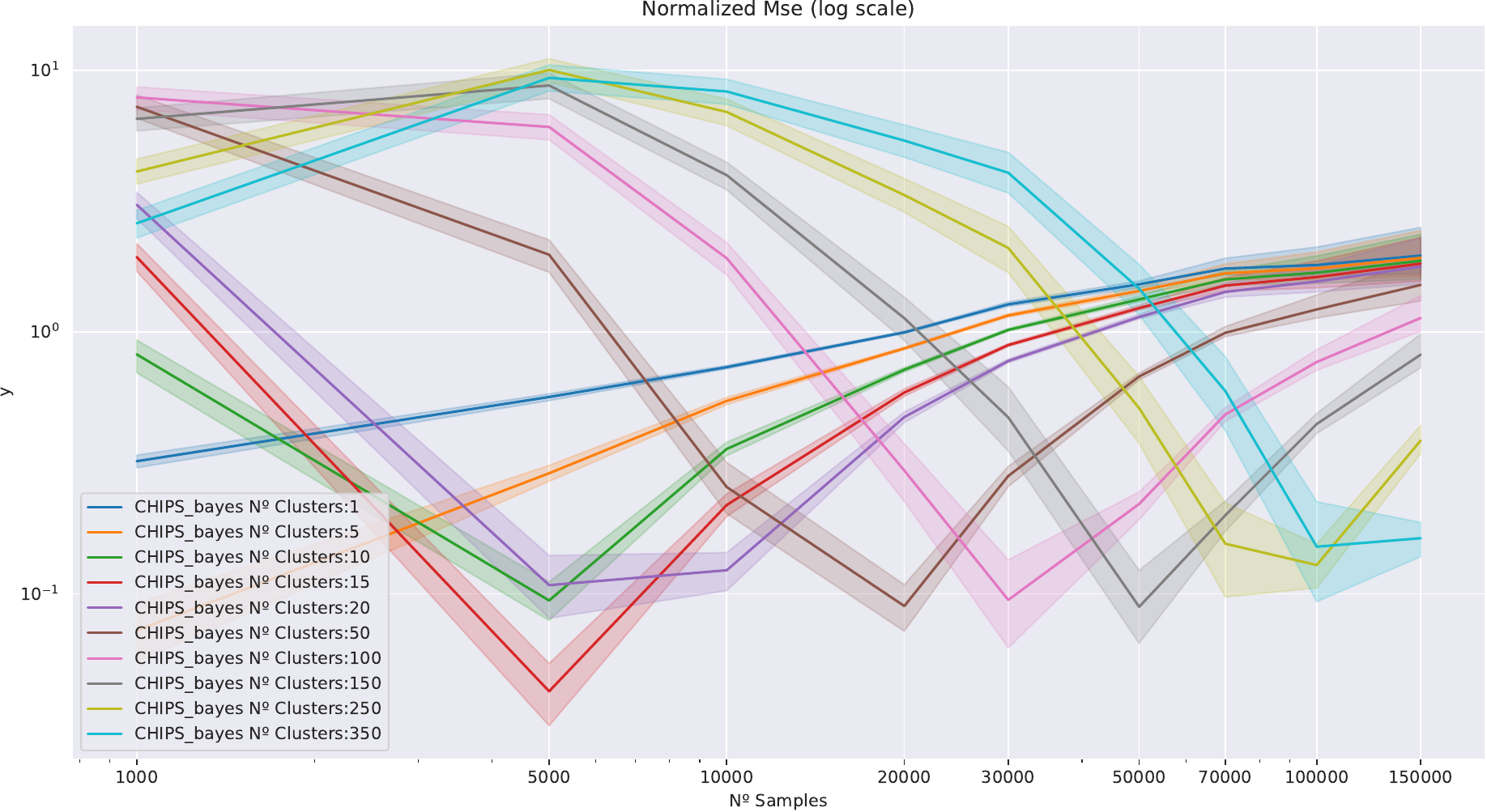} }}%
    \caption{Bi parametric experiments results using different number of clusters for analyzing CHIPS capabilities when increasing actions (a) and logging samples (b).}
    \label{fig:bivariate}%
\end{figure}

The experiments varying single parameters described in the previous section indicate that increasing the number of actions in a fixed configuration progressively deteriorates CHIPS’ performance. This behaviour is expected since the larger the action space, the more likely it is to incur in a situation in which \cref{as2} does not hold with a fixed number of clusters. In this situation, we found that reducing the number of clusters can mitigate the performance decay by
pooling information from broader contexts clusters while increasing it could be beneficial in reduced action spaces (see
\cref{fig:bivariate} (a)). Similarly, the number of samples from the logging policy also conditions how significant the performance gap between CHIPS and IPS is. In particular, the higher the number of samples, the more beneficial it is to use a higher number of clusters to try to obtain a more detailed partition structure of the context space, while a reduced number of clusters has an edge on few-sample cases (see Figure \ref{fig:bivariate} (b)).

\begin{figure}[ht]
    \centering
    \includegraphics[width=\textwidth]{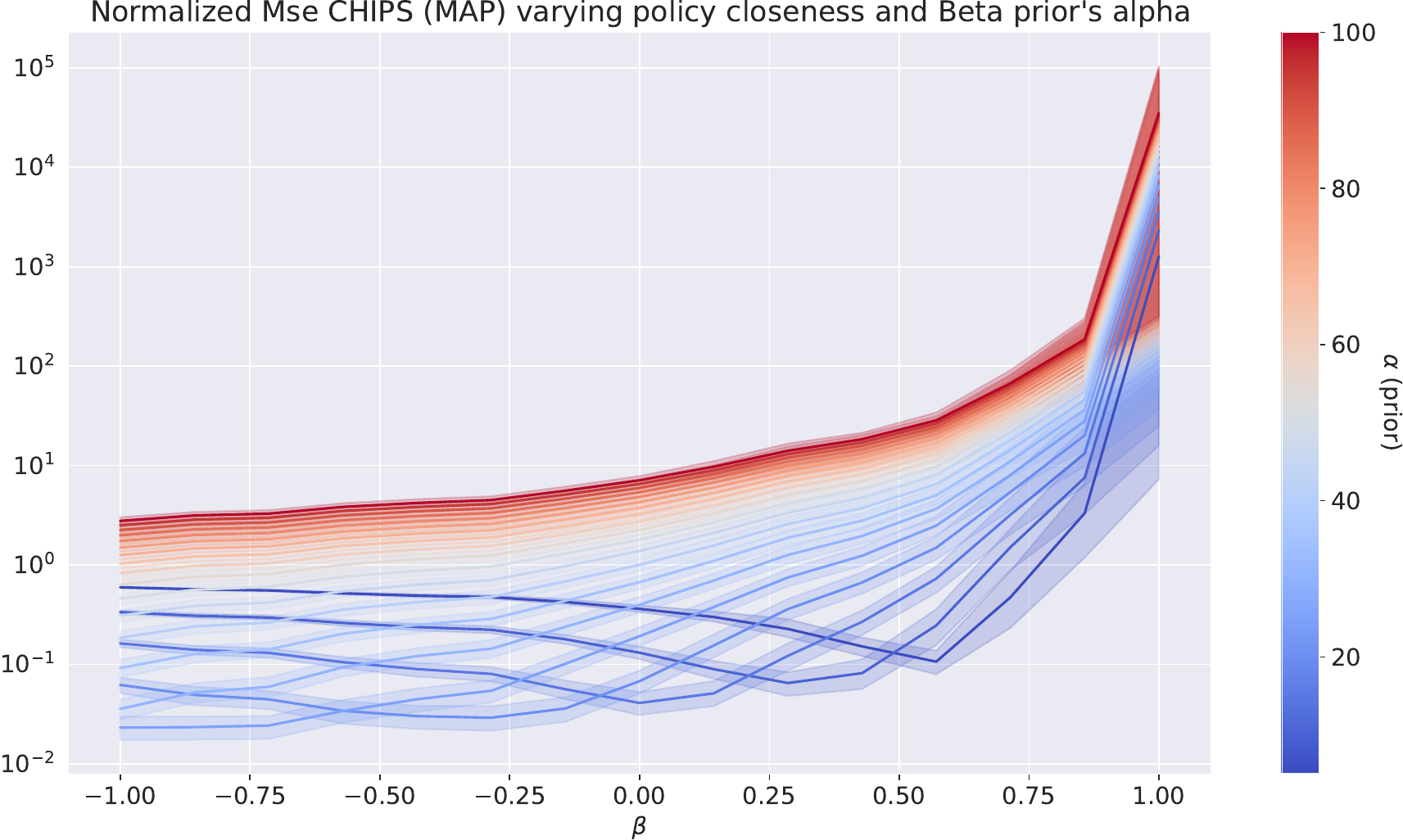}
    \caption{Normalized performance of CHIPS (MAP) with respect to IPS using different values for the $\alpha$ parameter in the Beta prior and distributional shift between logging and evaluation policies ($\beta$).}
    \label{fig:alpha_beta}
\end{figure}

We also study the effect of varying the $\alpha$ parameter in CHIPS' (MAP) Beta prior, using different values of the distributional shift between policies ($\beta$). In \cref{fig:alpha_beta}, we observe that mid values of $\alpha$ (30-50) offer better performance when there is a considerable distributional shift between logging and evaluation policy (i.e., $\beta \approx -1$) since the expected reward per cluster is pushed towards the prior's expectation, creating some resistance from converging to the average observed rewards (i.e., mitigating the reward misspecification existing under this conditions). As the distributional gap closes, lower values of $\alpha$ are more favourable since the samples observed per cluster are better representatives of the real expected reward. However, higher values for $\alpha$ (80-100) result in excessive resistance that deteriorates CHIPS' performance. It is also worth mentioning that as the distributional gap closes, CHIPS (MAP) loses its advantage with respect to IPS since the logging and evaluation policies are closer, and the ML estimates would offer better results, as previously shown in \cref{fig:syn}.

\subsection{Real Experiments}
\label{apdx:extra-real}
\begin{figure}[ht]%
    \centering
    \subfloat[\centering ECDFs of CHIPS (MAP) using different number of clusters in the real dataset.]{{\includegraphics[width=0.6\textwidth]{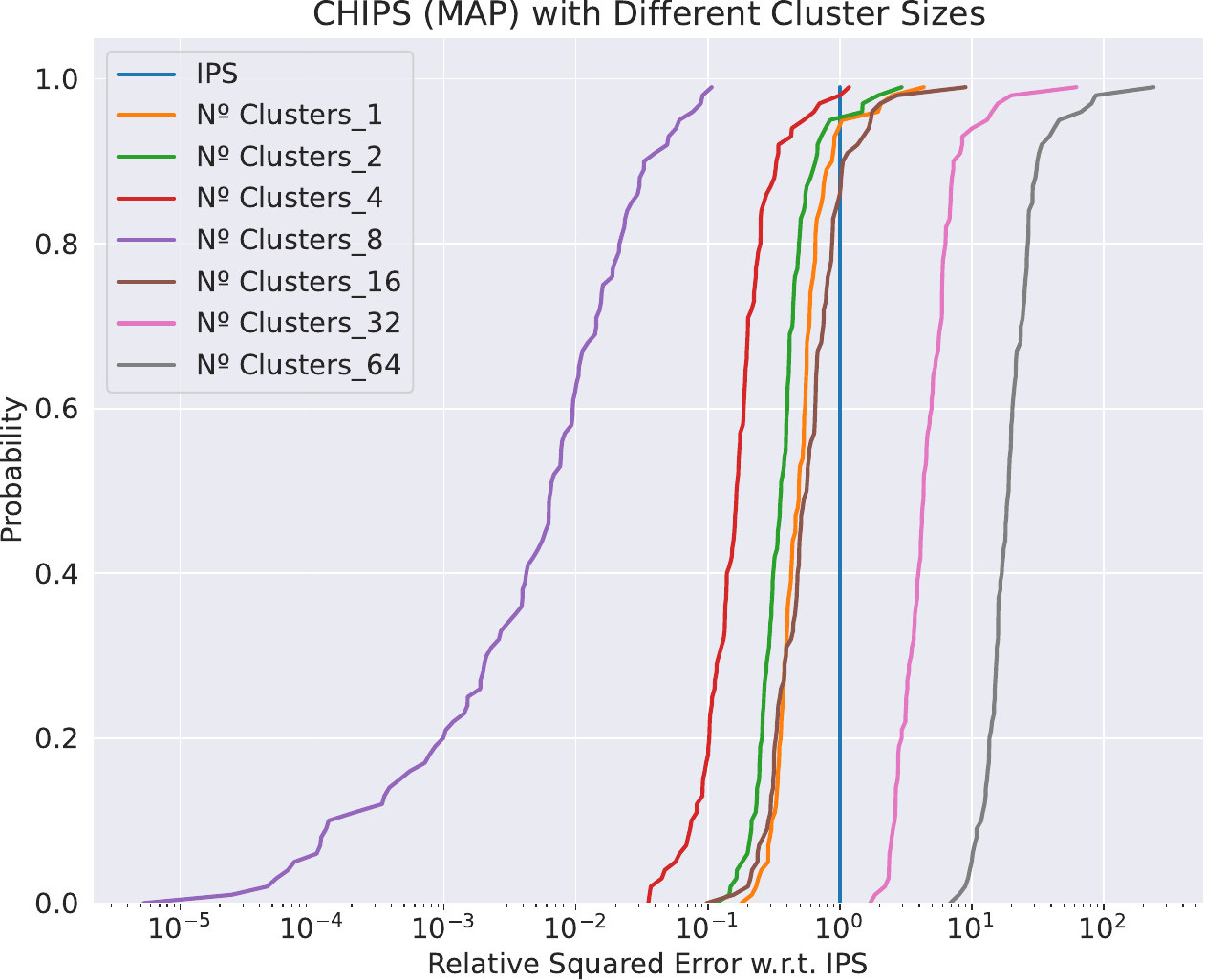} }}%
    \qquad
    \subfloat[\centering ECDFs of CHIPS (MAP) using different values of $\alpha$ for the Beta prior in the real dataset.]{{\includegraphics[width=0.6\textwidth]{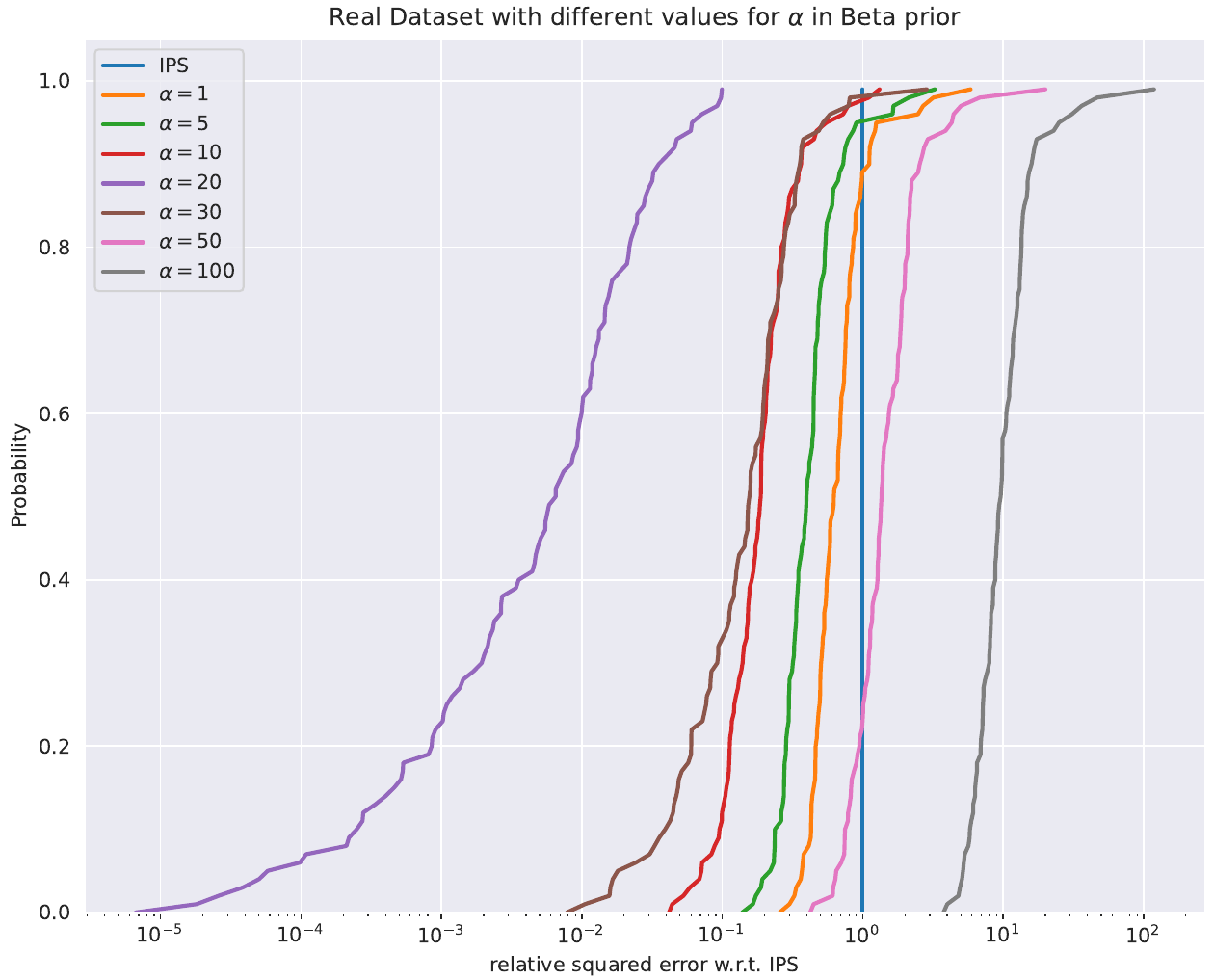} }}%
    \caption{Additional experiments varying the number of clusters and the $\alpha$ parameter in the Beta prior for CHIPS (MAP) in the real dataset (using 100000 samples).}
    \label{fig:real-extra}%
\end{figure}
\subsection{MAP vs ML}
\label{apdx:map-v-ml}
\begin{figure}[ht]
    \centering
    \includegraphics[width=\textwidth]{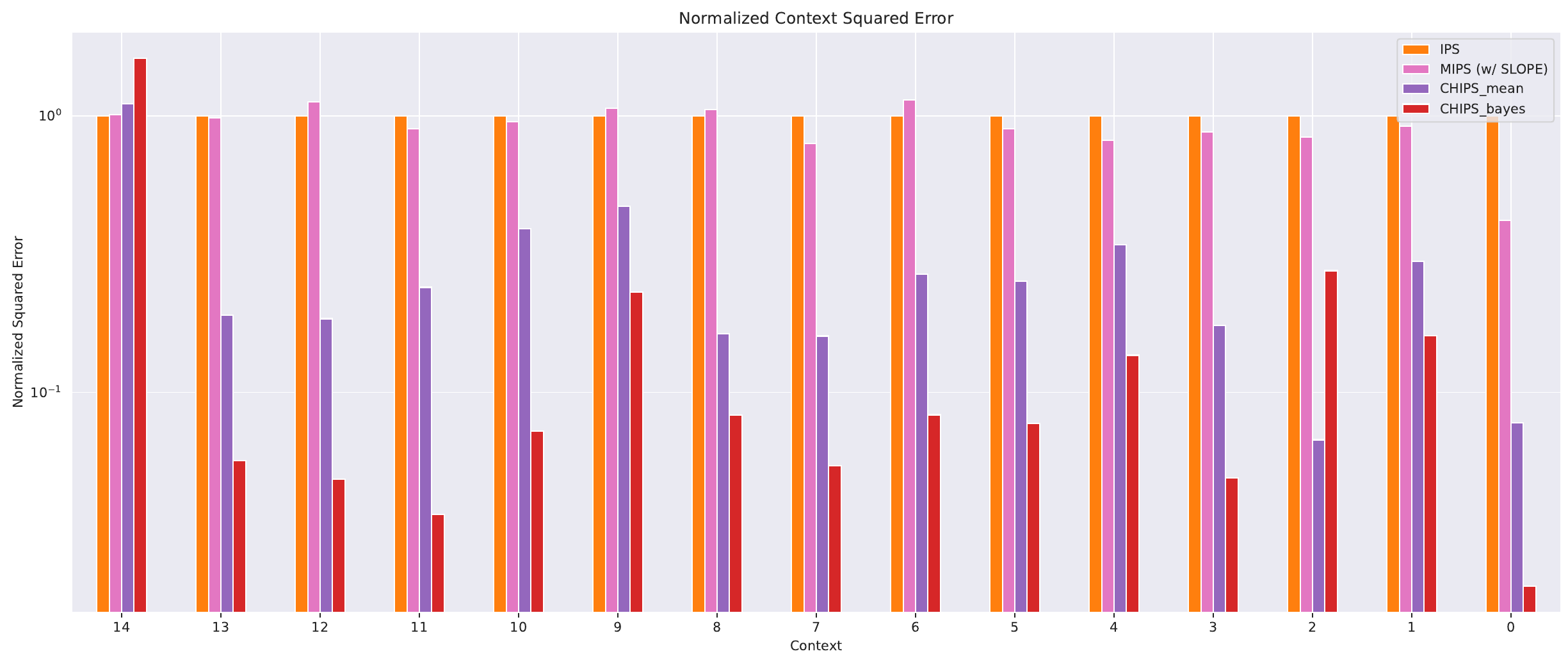}
    \caption{Normalized MSE with respect to IPS of the expected rewards for the 15 most common context-action pairs in the real logging dataset.}
    \label{fig:extra-weights-real}
\end{figure}

\begin{figure}[ht]
    \centering
    \includegraphics[width=\textwidth]{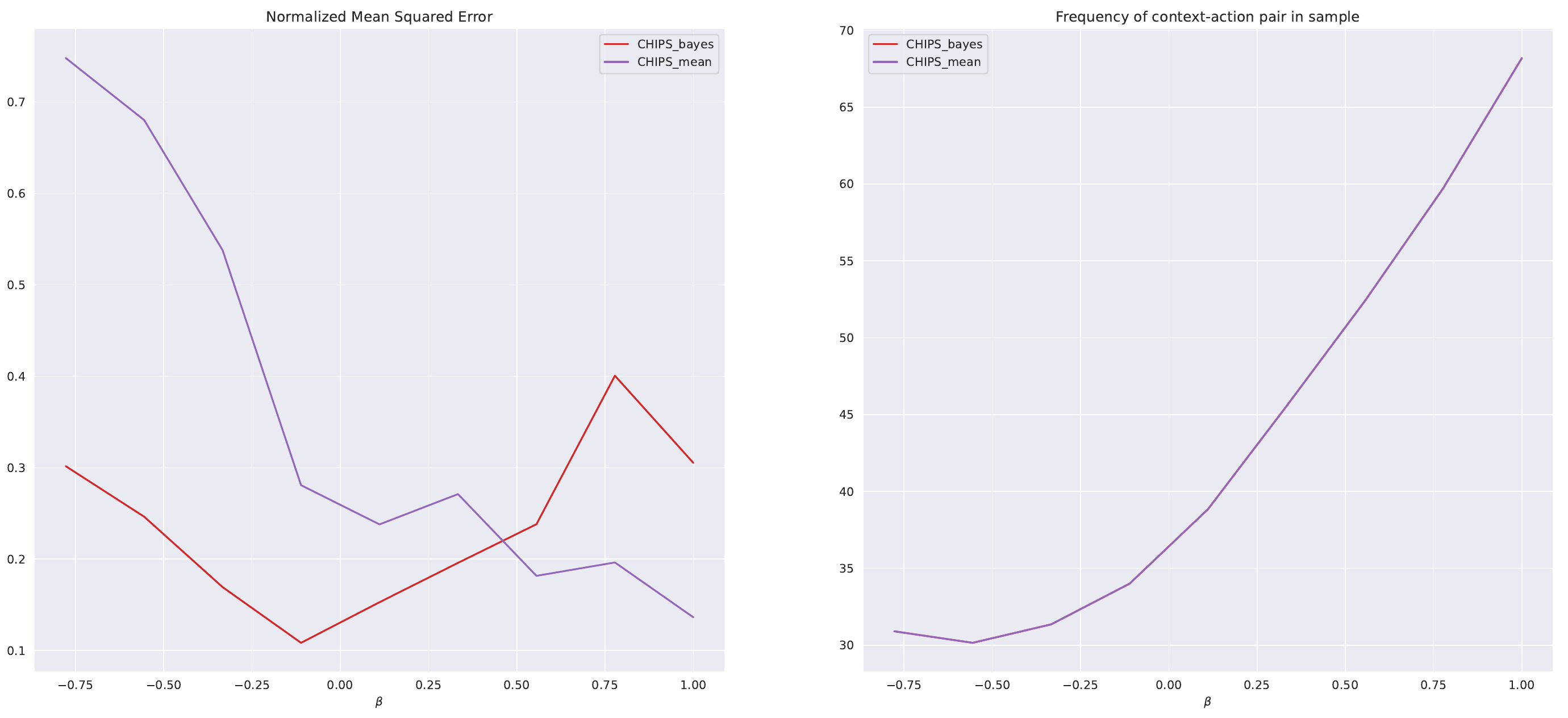}
    \caption{Normalized MSE of CHIPS with respect to IPS (left) and samples in the associated cluster (right) for the most common context-action pair in the evaluation policy while varying the distributional shift ($\beta$) in the synthetic dataset.}
    \label{fig:extra-weights-syn}
\end{figure}

In this section, we analyze the reason behind the jump in performance using the CHIPS estimator with the MAP estimate for
the expected reward per cluster. For this purpose, we have conducted two experiments, one in the synthetic dataset and the other in the real dataset. For the synthetic experiment, given a distributional shift value $\beta$, we select the most relevant context-action pair $(x^*,a^*)$ under the evaluation policy $\pi$ (i.e., $(x^*,a^*) = \arg\max_{(x,a) \in \X \times \A} \pi(a|x)$). Then we analyze the mean squared error of the expected reward (given $x^*$) estimations made by CHIPS (MAP) (i.e., $w(a^*,c^*)\hat{r}_\text{bayes}(a^*,c^*)$) and CHIPS (ML) (i.e., $w(a^*,c^*)\hat{r}_\text{mean}(a^*,c^*)$) w.r.t IPS (where $c^*$ is the cluster associated with $x^*$). We also compute the number of observations in $c^*$ in which action $a^*$ was selected. This process is repeated 100 times with different policies generated under different random seeds, and the results for the number of samples per cluster and squared errors are averaged. We repeat this for ten different values of $\beta$ ranging from -1 to 1 and represent the moving averages for relative squared errors and samples in \cref{fig:extra-weights-syn}. We observe that the number of samples per cluster increases with $\beta$ as both policies become closer. This increase in the number of samples makes the ML estimates progressively more accurate since the extra samples push the estimated expected value to the real expected value. For lower values of $\beta$, when the gap between policies is more significant, although some samples are available in the cluster, the values for the rewards observed on them are non-informative of the real expected value (hence the difficulty of ML to make an accurate estimation and the difference between MAP and ML for misspecified reward settings as depicted in \cref{fig:syn}).
For the real dataset, we follow a similar procedure, but instead of the most relevant context-action pair, we select the top 15 and compare the conditional expected reward estimates MSE with respect to IPS' (see \cref{fig:extra-weights-real}). Since the logging policy for this dataset is uniform, the distributional shift between the logging and evaluation policies is not as significant as the one presented in the base configuration of the synthetic dataset ($\beta=-1$). In practice, this means that the CHIPS estimation of the expected reward per cluster using ML is more accurate than in the synthetic dataset but still far from the performance jump of the CHIPS estimate using MAP, as we would expect from the results in \cref{fig:real}.

\subsection{Choosing alpha in Arbitrary Problems}
\label{apdx:alphas}
\begin{figure}[ht]
    \centering
    \includegraphics[width=\textwidth]{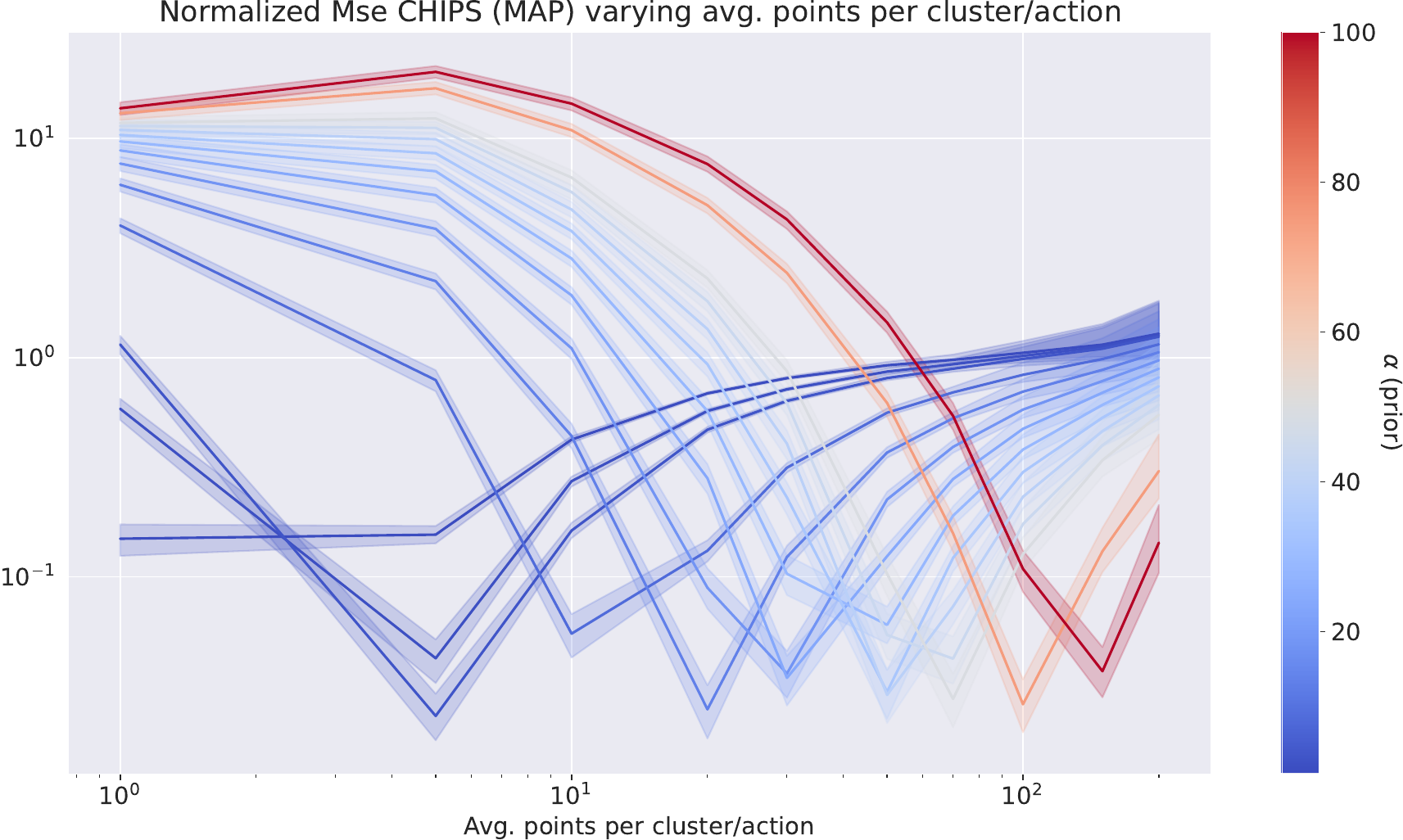}
    \caption{Normalized MSE of CHIPS (MAP) with respect to IPS using different values of $\alpha$ and number of expected data points per cluster-action.}
    \label{fig:avg-data-alpha}
\end{figure}

In \cref{fig:real-extra} (b), we observe that the hyperparameters of the MAP estimation process can heavily impact the performance of the method. As previously discussed in \cref{apdx:bi}, MAP hyperparameters control the resistance with which the expected reward per cluster is \textit{pulled} towards the prior’s expectation. This resistance is particularly noticeable in smaller size clusters, in which estimating a reward based on observations alone is much more challenging. Since in these clusters the partitioning method cannot ensure high homogeneity at reward level, in our experimentation we decided to use a non-informative prior (i.e., $\alpha = \hat{\beta}$), to mitigate possible violations of \cref{as3} and reward misspecification. Intuitively, an optimal value for $\alpha$ under these conditions needs to balance the prior's resistance to prevent reward misspecification without incurring into creating a quasi-uniform reward estimation (excessively large values of $\alpha$). In \cref{fig:avg-data-alpha} we explore the optimal value of $\alpha$ for a given average number of datapoints per cluster-action. As expected, for small size clusters, lower values of $alpha$ are favoured since the pull towards the prior's expectation is soft, while on bigger clusters, the value of $\alpha$ (and consequently the resistance) needs to grow to effectively control reward misspecification (otherwise the expected reward value would be pulled towards the value of the observed samples).

\begin{figure}[ht]
    \centering
    \includegraphics[width=\textwidth]{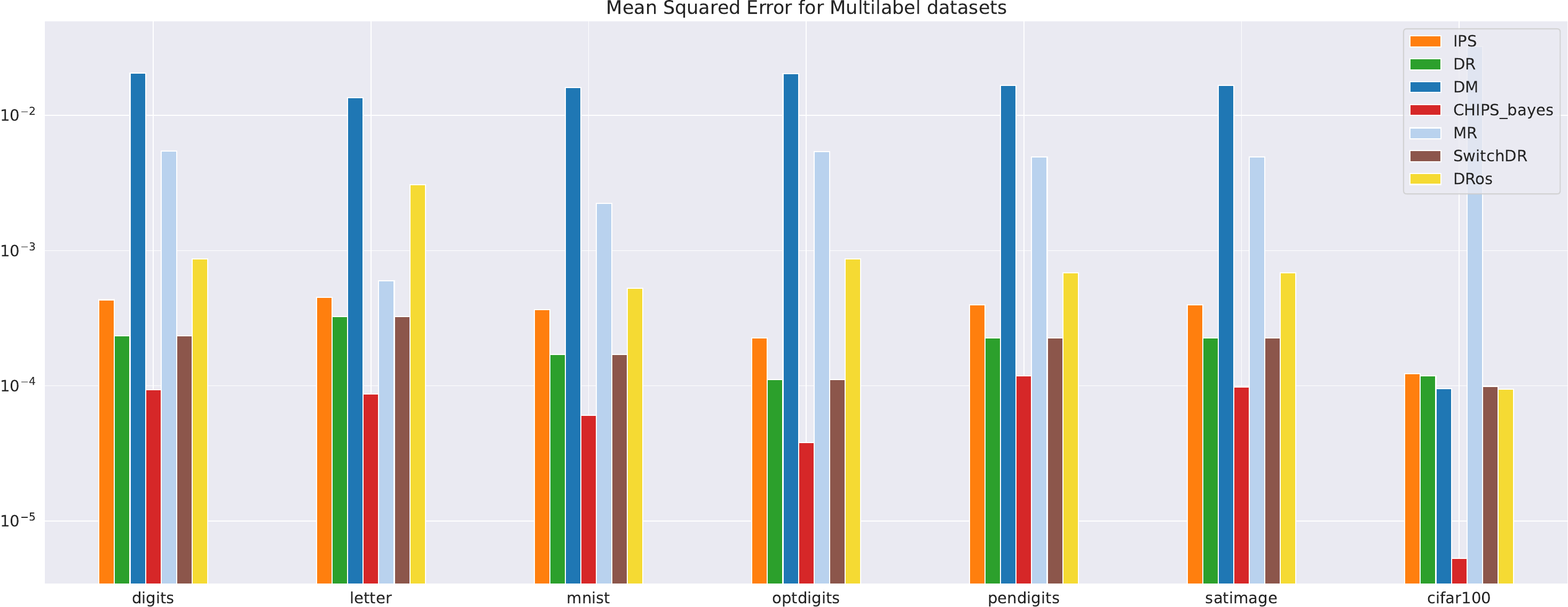}
    \caption{MSE of CHIPS (MAP) using $\alpha$ selection policy with respect to IPS, DR, DM, MR \protect\citep{taufiq2023marginal}, DRos\protect\citep{dros} and SwitchDR \protect\citep{switch}.}
    \label{fig:multiclass}
\end{figure}

To choose the value of $\alpha$ in an arbitrary problem, we propose the following selection process:
\begin{enumerate}
    \item Determine the number of clusters to use depending on the number of clusters (reference in \cref{fig:bivariate} (a).
    \item Partition the context space $\X$ in clusters ${c_1, c_2 ,..., c_n}$.
    \item Generate synthetic data $\hat{X_{\text{ev}}}$ using $\X_{\text{train}}$ and $\pi_e$.
    \item Estimate number of average data points per cluster-action from $\hat{X_{\text{ev}}}$.
    \item Choose $\alpha$ from the reference in \cref{fig:avg-data-alpha}.
\end{enumerate}
For testing this selection process, following the experimental protocol of \citet{taufiq2023marginal}, we transform five UCI datasets \citep{Dua:2019}, MNIST \citep{mnist}
, and CIFAR-100 \citep{cifar100} from multi-class classification problems into contextual bandits data \citep{dr}. The results (averaged 50 times) in \cref{fig:multiclass} show a consistent improvement with respect to existing methods, empirically proving the effectiveness of the $\alpha$ selection process.

\begin{figure}[ht]
    \centering
    \includegraphics[width=\textwidth]{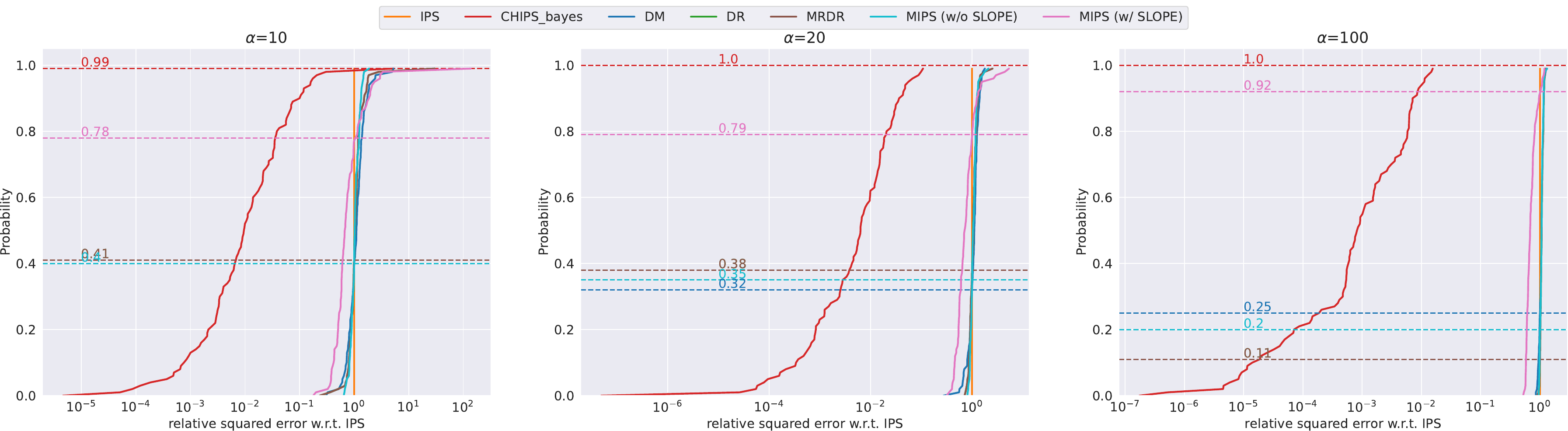}
    \caption{ECDF of the relative mean squared error with respect to IPS for the real dataset using 50000 (left), 100000 (center), and 500000 (right) logging samples and the $\alpha$ selection process.}
    \label{fig:real-3-extra}
\end{figure}

Additionally, we perform an alternative experiment using the real dataset, in which instead of fixing $\alpha$ and vary the number of clusters according to the reference in \cref{fig:bivariate} (b) with 50000, 100000 and 500000 samples (see \cref{fig:real}, we follow the $\alpha$ selection process, fix the number of clusters and increase the value of $\alpha$ according to \cref{fig:avg-data-alpha}. In \cref{fig:real-3-extra} we observe equivalent results as in our previous experiment confirming the equivalence of using a reference for the number of samples and varying the number of clusters with a fixed value for $\alpha$, or varying $\alpha$ with a fixed number of clusters obtained by using a reference for the number of actions.
\section{CLUSTER STRUCTURE IN DATASETS}
\begin{figure}[ht]%
    \centering
    \subfloat[\centering Expected reward per context for a sepcific action in the synthetic dataset.]{{\includegraphics[width=\textwidth]{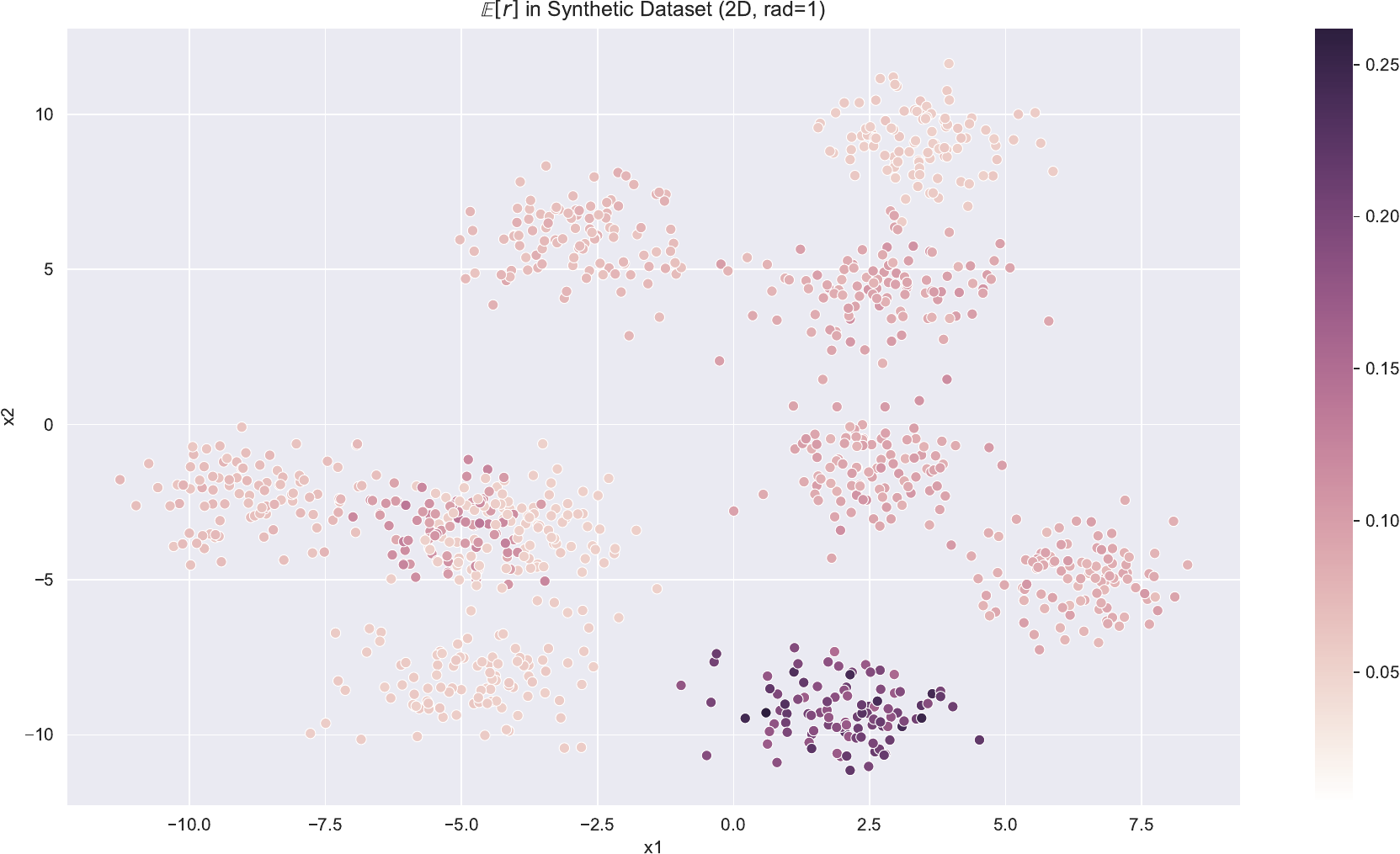} }}%
    \qquad
    \subfloat[\centering Action maximizing the expected reward per context in the synthetic dataset.]{{\includegraphics[width=\textwidth]{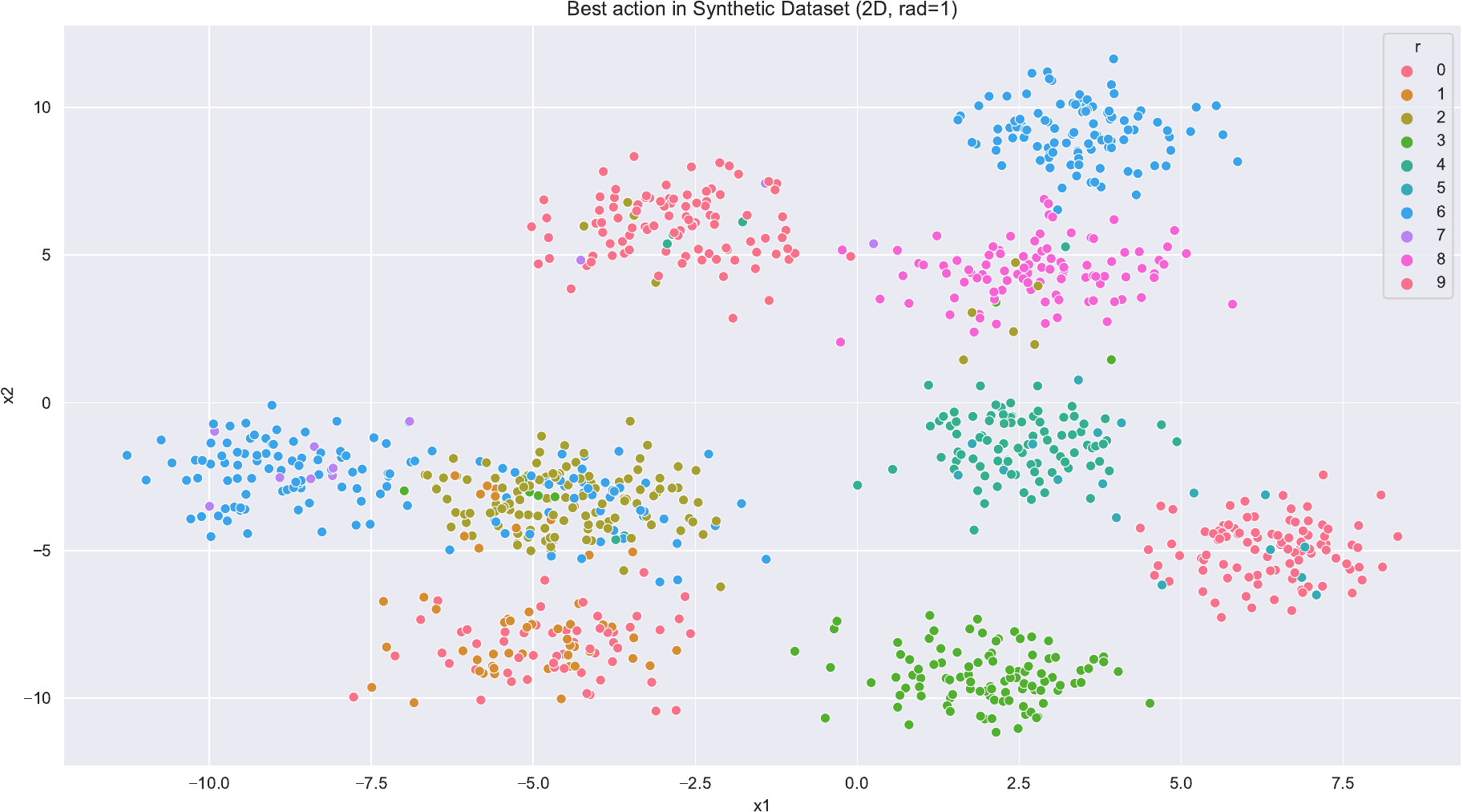} }}%
    \caption{Representation of the synthetic dataset using 2-dimensional contexts.}
    \label{fig:synthetic-dataset}% 
\end{figure}

The generated synthetic dataset ensures that the expected reward inside a cluster is similar and that the best possible action is usually the same for all the context within the cluster (see \cref{fig:synthetic-dataset}), mimicking real-world settings like e-commerce in which we can expect similar behaviour for close contexts.

\section{EXPERIMENTAL PROTOCOL}
\label{apdx:experimental-protocol}
For evaluation in the real dataset, we follow \citep{mips} protocol to evaluate estimators' accuracy given two sources of data. Given a logging policy $\pi$, a dataset collected under it $\D$, a logging policy $\pi_0$, and the dataset collected under it $\D_0$, we follow the following procedure:
\begin{enumerate}
    \item Extract $n$ independent bootstrap samples with replacement from the logging dataset $\D_0^* := \{(x_i, a_i, r_i)\}_{i=1}^n$.
    \item Estimate the policy value of $\pi$ using the sample $\D_0^*$. We denote this estimate as $\hat{V}(\pi; \D_0^*)$.
    \item Compute the relative mean squared error with respect to IPS: 
    \begin{equation*}
    \mathcal{Z}(\hat{V}, \D_0^*) = \frac{\left(V(\pi) - \hat{V}(\pi; \D_0^*)\right)^2}{\left(V(\pi) - \hat{V}_{\text{IPS}}(\pi; \D_0^*)\right)^2}
    \end{equation*}
    Where $V(\pi) := \frac{1}{|\mathcal{D}|} \sum_{(\cdot,\cdot,r_i) \in \D} r_i$.
    \item Repeat steps 1,2, and 3 $T=100$ times and compute the Empirical Cumulative Distribution Function (ECDF) as:
    \begin{equation*}
    \hat{F}_{\mathcal{Z}}(x) := \frac{1}{T}\sum_{t=1}^T\mathbb{1}\{\mathcal{Z}_t(\hat{V}, \D_0^*) \leq x\}
    \end{equation*}
\end{enumerate}
\section{COMPLEXITY}
\label{appdx:time}
\begin{figure}[ht]%
    \centering
    \subfloat[\centering Execution times for the real dataset including DM, DR, and MRDR.]{{\includegraphics[width=\textwidth]{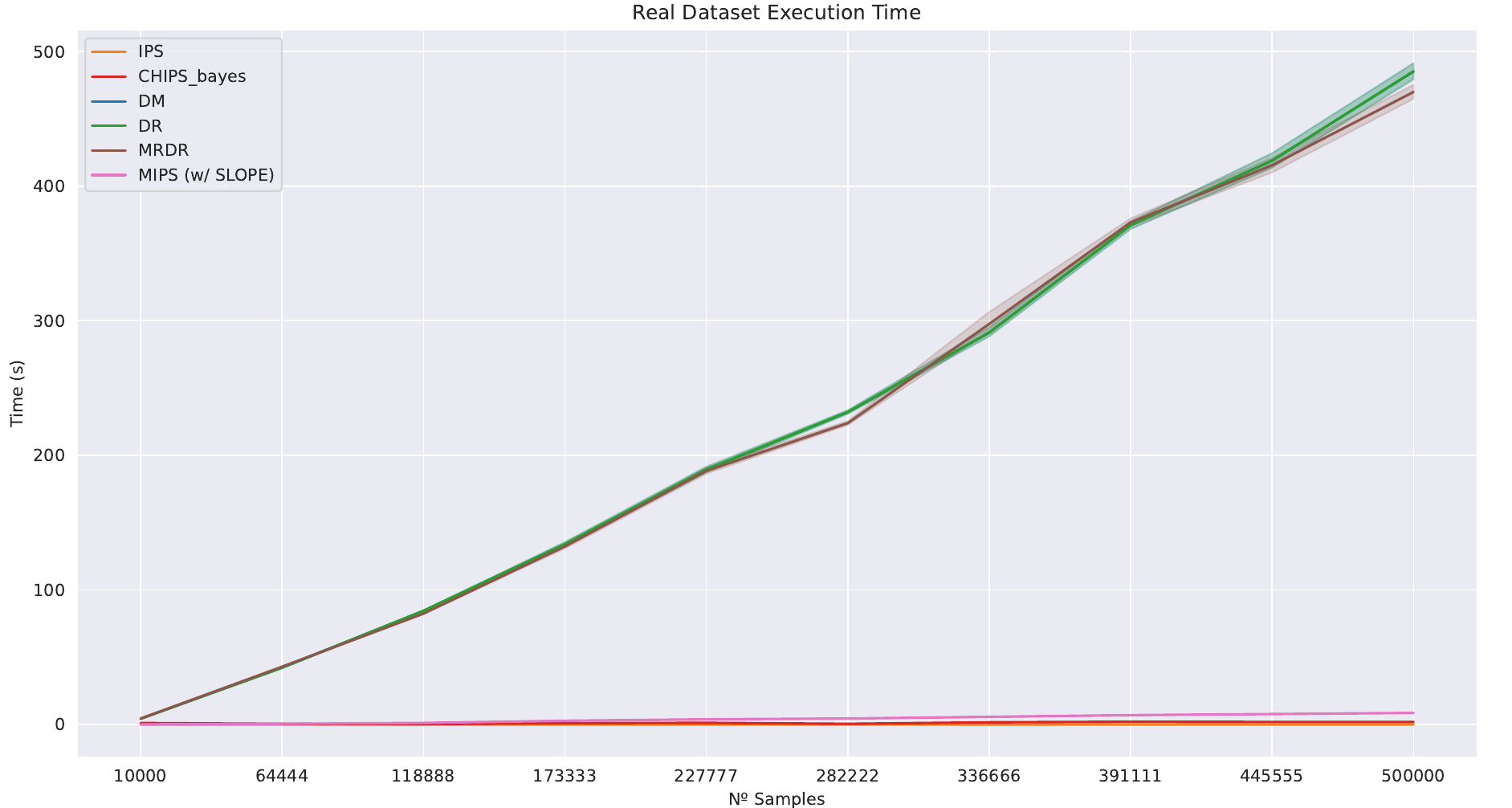} }}%
    \qquad
    \subfloat[\centering Execution times for the real dataset for IPS, CHIPS, and MIPS.]{{\includegraphics[width=\textwidth]{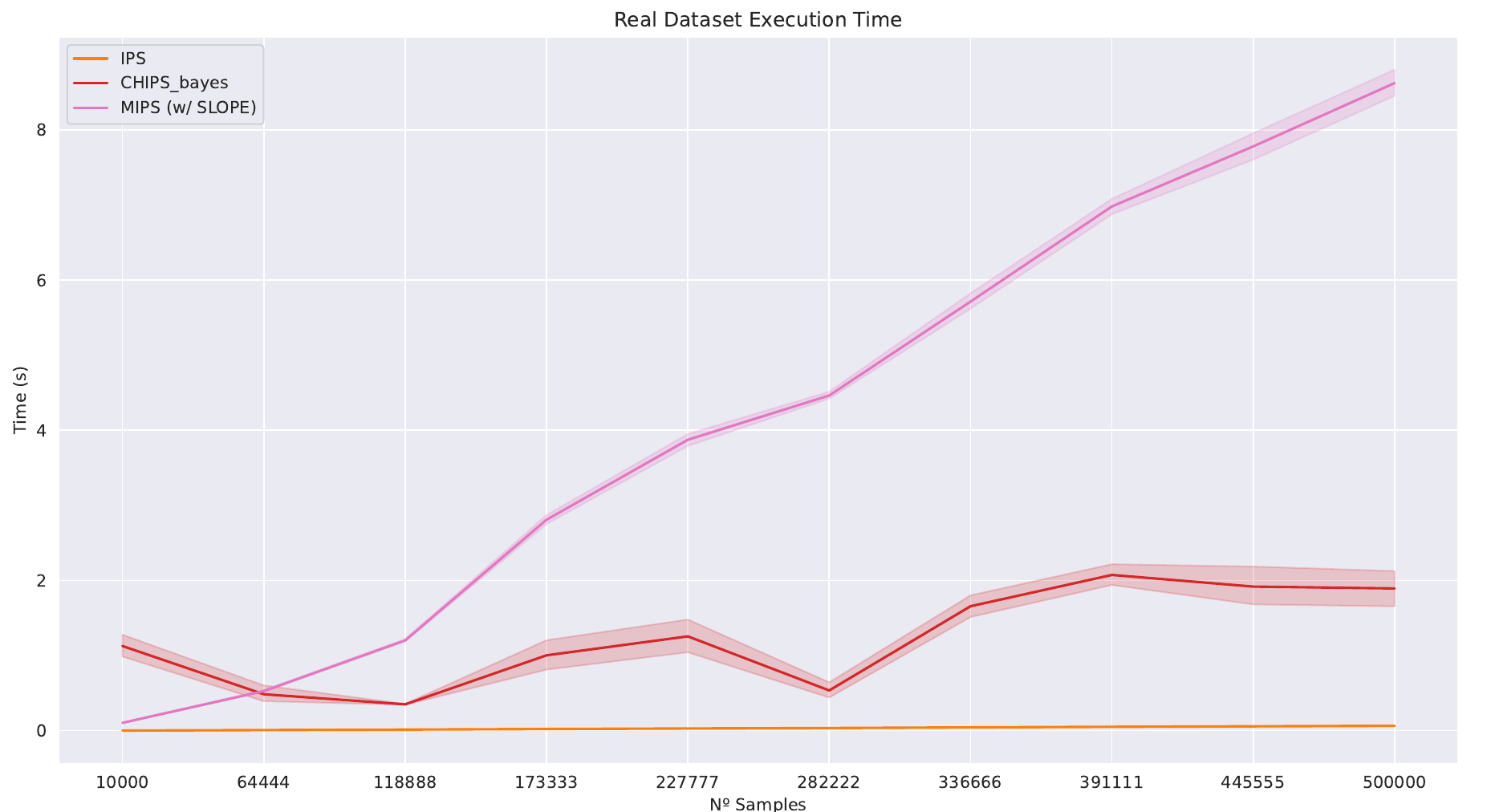} }}%
    \caption{Average execution times increasing the sample size (100 executions per sample size).}
    \label{fig:times}%
\end{figure}

Algorithmically, since the CHIPS estimator can be regarded as performing the same procedure as IPS with different weights and rewards,the time complexity given $n$ logging samples, and clustering method $\xi$ can be expressed as:
$$
\text{complexity}(\text{CHIPS}(n; \xi)) = \text{complexity}(\text{IPS}(n)) + \text{complexity}(\xi(n))
$$
For example, since the time complexity of IPS is $\mathcal{O}(n)$, using DBSCAN ($\mathcal{O}(n\log n)$) as a clustering method, we would get a time complexity for CHIPS of $\mathcal{O}(n\log n)$. In our experiments, we used batch-Kmeans \citep{kmeans} as clustering method, that has a time complexity of $\mathcal{O}(mkd_xt)$ where $m$ is the batch size, $k$ is the number of clusters, $d_x$ is the dimension of the features and $t$ is the number of iterations.
In the implementation used, we fixed $m=1024$ and $t=100$, therefore, in this case, the time complexity of the CHIPS method is $\mathcal{O}(kd_x) + \mathcal{O}(n)$. The time complexity of the MIPS estimator can be estimated similarly as $\mathcal{O}(nd_e) + \mathcal{O}(n) = \mathcal{O}(nd_e)$, where the $\mathcal{O}(nd_e)$ term comes from the logistic regression used to estimate $\pi_0(a|x,e)$ (being $e$ an action embedding) and $d_e$ is the action embedding dimension. The methods using a supervised classifier (DM, DR, and MRDR) get their dominant term in time complexity from the training process of the classifier, in our case $\mathcal{O}(n d s \log n )$ with $s$ being the number of trees. In practice, this means that DM, , and MRDR will have significantly higher execution times (see \cref{fig:times} (a)), and CHIPS will generally be faster than MIPS since $k << n$ to leverage the cluster structure, as we can appreciate in \cref{fig:times} (b). The space complexity of CHIPS can be estimated following a similar approach, for example when using the KMeans algorithm the space complexity is $\mathcal{O}(n(k+d)) + \mathcal{O}(n) = \mathcal{O}(n(k+d))$.

\end{document}